\documentclass[10pt,journal,compsoc]{IEEEtran}
\usepackage{times}
\usepackage{graphicx}
\usepackage{epsfig}
\usepackage{amsmath}
\usepackage{amssymb}
\usepackage{epstopdf}
\usepackage{amsmath}
\usepackage{makecell}
\usepackage{amsfonts}
\usepackage{color}
\usepackage{multirow}
\usepackage{setspace}
\usepackage{booktabs}
\usepackage{enumerate}
\usepackage{pifont}
\usepackage{ragged2e}
\usepackage{colortbl}
\usepackage[linesnumbered,ruled,vlined]{algorithm2e}
 
\usepackage{subfigure}
\definecolor{mygray}{gray}{.95}
\ifCLASSOPTIONcompsoc
  \usepackage[nocompress]{cite}
\else
  \usepackage{cite}
\fi

\ifCLASSINFOpdf
\else
\fi

\begin{document}

%\title{Complementary-View Multiple Human Association and Tracking}
%\title{Multiple Human Association and Tracking from Complementary-View Cameras}
\title{Unveiling the Power of Self-supervision for Multi-view Multi-human Association and Tracking}
\author{Wei~Feng,~\IEEEmembership{Member,~IEEE,}
        Feifan~Wang,
        Ruize~Han, \IEEEmembership{Student Member,~IEEE,}
        Zekun~Qian,
        and~Song~Wang,~\IEEEmembership{Senior~Member,~IEEE}% <-this % stops a space
\IEEEcompsocitemizethanks{\IEEEcompsocthanksitem  W.~Feng, F. Wang, R. Han (Corresponding Author) and Z. Qian are with the College of Intelligence and Computing, Tianjin University, Tianjin, 300350, China, and with the Key Research Center for Surface Inspection and Analysis of Cultural Relics, SACH, China.\protect\\
S.~Wang is with the Department of Computer Science and Engineering, University of South Carolina, Columbia, SC 29208, USA.\protect\\
%E-mail: wfeng@ieee.org (W.~Feng); songwang@cec.sc.edu (S.~Wang).

%\IEEEcompsocthanksitem $^\dagger$R.~Han is the corresponding author.
}% <-this % stops an unwanted space
}

%\markboth{IEEE TRANSACTIONS ON PATTERN ANALYSIS AND MACHINE INTELLIGENCE,~IN SUBMISSION.}%
%{R.~Han \MakeLowercase{\textit{et al.}}: Multiple Human Association and Tracking from Egocentric and Complementary Top Views}


\IEEEtitleabstractindextext{%
	
\begin{abstract}
\justifying
Multi-view {multi}-human association and tracking (MvMHAT), {is} a new but important problem for multi-person scene video surveillance, aiming to track a group of people over time in each view, as well as to identify the same person across different views at the same time, which is different from previous MOT and  multi-camera MOT tasks only considering the over-time human tracking.
%This is a relatively new problem but is very important for multi-person scene video surveillance. 
{This way, the videos for MvMHAT require more complex annotations while containing more information for self learning.} In this work, we tackle this problem with a self-supervised learning aware end-to-end network. 
%to {learn the appearance feature representation and spatial-temporal association results based on the same insight.}
Specifically, we propose to take advantage of the spatial-temporal self-consistency rationale by considering three properties of {reflexivity}, symmetry and transitivity.
Besides the reflexivity property that naturally holds, we design the self-supervised learning losses based on the properties of symmetry and transitivity, for both appearance feature learning and assignment matrix optimization, to associate the multiple humans over time and across views.
Furthermore, to promote the research on MvMHAT, we build two new large-scale benchmarks for the network training and testing of different algorithms. 
Extensive experiments on the proposed benchmarks verify the effectiveness of our method. 
We have released the benchmark and code to the public.

\end{abstract}

}

\maketitle

\IEEEdisplaynontitleabstractindextext

% \fi
\IEEEpeerreviewmaketitle

\IEEEraisesectionheading{\section{Introduction}\label{sec:introduction}}
%\section{Introduction}\label{sec:introduction}

% Background
\IEEEPARstart{M}{ultiple}
 object tracking (MOT), especially multiple human tracking, is a fundamental and important task in computer vision~\cite{berclaz2011multiple, sun2019deep, chu2019famnet}. 
 In this paper, we study the problem of multi-view multi-human association and tracking (MvMHAT), an extension of MOT, which aims to continuously track a group of people in each view while simultaneously identifying the same persons {across} multiple views at each time~\cite{han2021multiple}.	
 With MvMHAT, we can not only record the temporal trajectories of the 	involved humans (referred to as subjects in this paper), but also comprehensively observe the subjects' details, e.g., the human pose and behavior, {by combining information from} different views, which {are desirable in} many potential real-world applications. 	
 A typical example is video surveillance -- imagine a scenario with multiple {pre-installed} or wearable cameras {covering a scene of multiple subjects} from different views, we can associate and analyze the collected videos for collaborative human activity recognition, important/abnormal person detection, etc., {based on the MvMHAT results.} 
 
\begin{figure}[t]	\vspace{-0cm}
	\centering
	\includegraphics[width=1\linewidth]{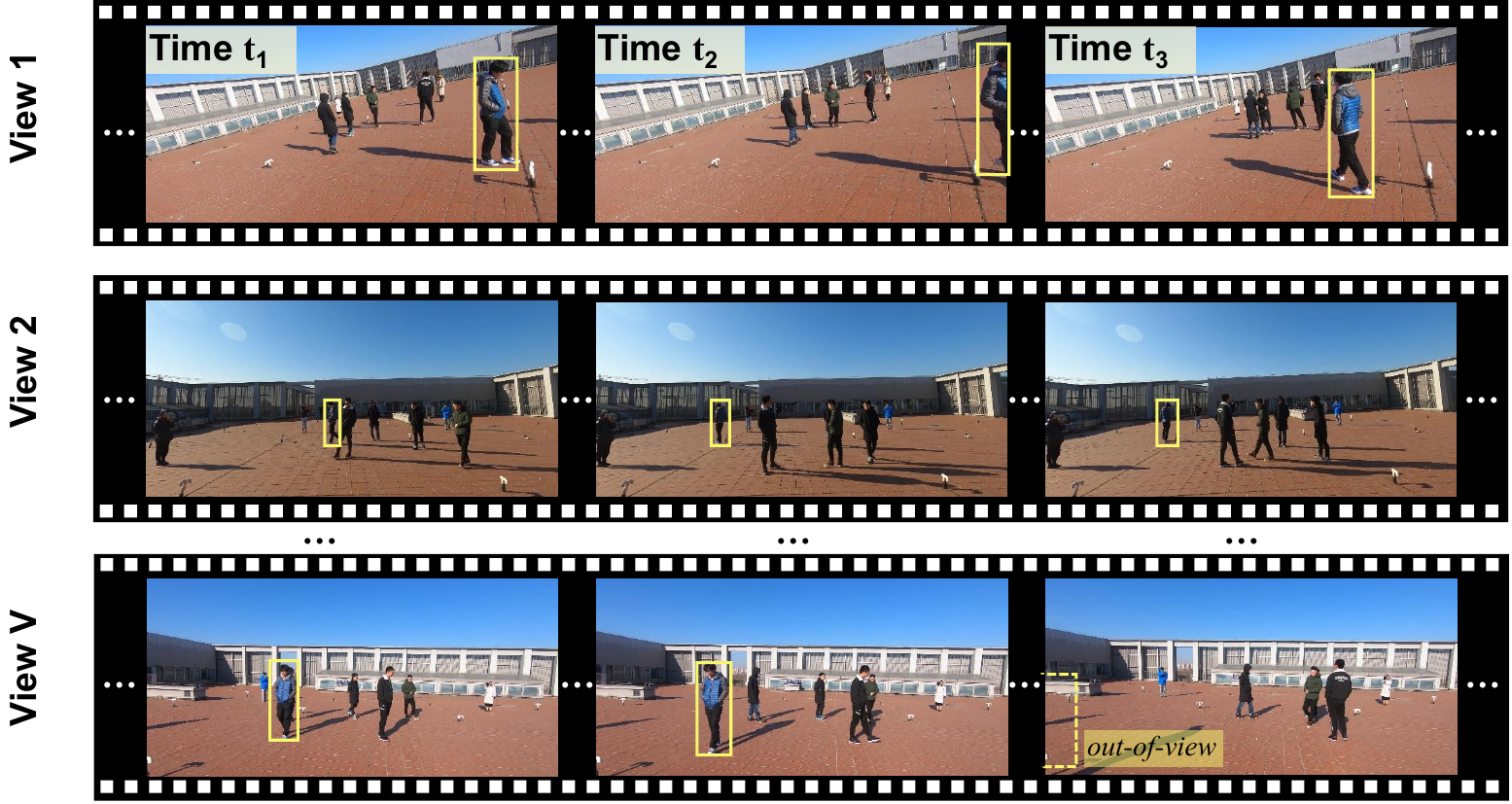}%\vspace{-10pt}
	\caption{An illustration of the proposed MvMHAT problem.}
	\label{fig:example} 
%	\vspace{-15pt}
\end{figure}

 Compared to MOT, MvMHAT is a more challenging problem since we need to associate all the subjects appearing in different views during the 
 tracking. The association also suffers from the unknown and large view differences, illumination differences, e.g., View \#1 and View \#2 in 
 Fig.~\ref{fig:example}, and the {difference} of contained subjects, e.g., Views \#1 and \#V at Time $t_3$, etc.
 Moreover, with the \textit{uncalibrated} multi-view cameras, e.g., wearable cameras, many existing multi-view human association methods are not {applicable} without the  required camera relative pose as input~\cite{3Dposetracking,Ass4Tracking}.
 Also, the human motion feature, a core cue for human matching in tracking, is usually inconsistent {across} 
 different views and may not be effective when used for measuring the subject similarity for the cross-view association. 
 {In this case,} the subject appearance representation becomes particularly important.
 
 So far, MvMHAT is a relatively new task with a handful of 	studies~\cite{2008Multicamera,xu2017cross,han2021multiple}.	Among them, most mainly study the over-time human tracking but not {much in-depth study and evaluation on} the cross-view human association results~\cite{2008Multicamera,xu2017cross,3Dposetracking,Ass4Tracking}.	
 {Note that by focusing only} on the temporal tracking in respective view but not the correspondence {across} different views, the advantage of the multi-view multi-human tracking is not fully exploited.
 A couple of recent works~\cite{han2021multiple,Han2019Complementary} study the MvMHAT with two {specific} complementary views{, i.e., top and horizontal views.}
 In this paper, as shown in Fig.~\ref{fig:example}, we are interested in MvMHAT in a more general setting where (arbitrary) multiple cameras (without prior calibration) are used to observe a multi-person scene from different views.

% Motivation 
%% Challenge 

%% Solution
For this purpose, we propose to develop a unified framework for the MvMHAT with arbitrary number of views.
As discussed above, a key point for this problem is to learn the human appearance similarity and the association relations of the subjects among all views over time.
As we know, most previous works for MOT learn an appearance model from abundant labeled data for the over-time appearance measurement.
On one hand, {for MvMHAT, the labor cost of annotation is greatly increased, which limits the emergence of some large-scale and high-quality datasets for network training.}
On the other hand, in the MvMHAT problem, we actually have various appearance information of each subject along time and across views. For example, the same person appearing in a pair of frames (views or time) should present symmetric-consistent, i.e., given two frames, if one person in frame 1 has the highest similarity with someone in frame 2, then this person in frame 2 should also be the most similar to that person in frame 1; the same person appearing in multiple frames should also be transitive-consistent, i.e., given the matched person pairs appearing in frames 1\&2, and frames 2\&3, {her/his matching relation between frames 1\&3 should be the same as {the result of} associating these two matching pairs.} 
%and in multiple frames should be cycle-consistent. 
These two observations motivate us to adopt a self-supervised learning based method to utilize the spatial-temporal human appearance consistency {and association consistency} for MvMHAT.

% Detailed Method & Contribution
As discussed above, in this paper, we propose a self-supervised learning framework to solve the MvMHAT problem. 
In the training stage, our basic idea is to make use of the \textit{spatial-temporal self-consistency} of the humans in different frames (taken from different views or time) for \textit{both appearance feature and assignment matrix learning}.
Specifically, given several videos capturing a group of 
people from different views, we first sample several frames from 
different views and time, then we apply the convolutional neural networks 
(CNN) to learn the embedding feature of each subject. 
We use the feature similarity of the subjects in each pair of frames to generate the matching matrix. We then propose a couple of symmetric-consistency (SymC) and transitive-consistency (TrsC) as pretext task to supervise the matching matrix for learning the appearance feature in the self-supervised manner.
Then, we propose a spatial-temporal assignment network (STAN) to simultaneously model the over-time temporal association and the cross-view spatial association, which generates the assignment matrix considering the global structural information for MvMHAT.
The STAN is also trained using the SymC and TrsC alike losses in a self-supervised way.
In the inference stage, we leverage a new joint {association and tracking} scheme to solve the MvMHAT task.

Moreover, the current research on MvMHAT is restricted by the 
lack of an appropriate public dataset that can be accessed and used to 
train and evaluate the deep network based algorithms.
In this paper, we build two new large-scale benchmarks based on several 
public datasets and self-collected data for the training and testing of 
the MvMHAT algorithm. 
%Extensive experiments on the proposed datasets verify the effectiveness 
%of our method. 
% Summary
The main contributions of this paper are:\\
%\begin{itemize}
1) We propose a self-supervised learning framework for MvMHAT, which in-depth excavate the potential of spatial-temporal self-consistency rationale in this problem. To the best of our knowledge, this is the first work to model such a problem following a 
self-supervised manner.\\
2) We propose the pairwise symmetric-consistency and triplewise transitive-consistency pretext tasks, which are guaranteed by the theoretical support and effectively used for both appearance feature and assignment matrix learning in our method.
They are also modeled as differentiable loss functions, to build the end-to-end framework for the cross-view and over-time subject association and tracking.\\
3) We build two new benchmarks for training and testing MvMHAT. 
Extensive 
experiments on the proposed datasets verify the rationality of our problem definition, the usefulness of the proposed benchmark, and the 
effectiveness of our method. We have released the benchmark to the public\footnote{https://github.com/realgump/MvMHAT}.
%\end{itemize}
%\vspace{-10pt}

%This work is extended from our prior conference paper~\cite{gan2021self} with the major %improvements as below:\\
%\redcolor{This work is extended from our prior conference paper.}
The remainder of this paper is organized as below. 
Section 2 reviews the related works. 
Section 3 elaborates on the proposed approach in detail. 
Section 4 presents the benchmark used in this work. 
Section 5 provides the experimental results and analysis. After that, we provide a further discussion in Section 6 and a brief conclusion and future work in Section 7. 
This paper is a substantial extension from a preliminary conference version~\cite{gan2021self} with
a number of major changes. First, we add a new spatial-temporal assignment matrix learning module (Section~\ref{sec:assmatrix}), which shares the self-consistency rationale for the appearance feature learning module in~\cite{gan2021self} to together form a fully self-supervised end-to-end framework. 
Second, a new pseudo label generation strategy with dummy nodes used for more general MvMHAT cases is introduced in Section~\ref{sec:lossfeat}. Third, we include a new dataset {MMP-MvMHAT} and  significantly extend the experimental comparisons and analyses in Section~\ref{sec:experment}. 
Finally, we add the discussion about the limitations and future work in Sections~\ref{sec:discussion} and~\ref{sec:conclusion}.

\section{Related Work}

\textbf{MOT} is a classical problem and has 
many applications in video processing and analysis.
The most famous framework for MOT is a tracking-by-detection scheme, in 
which 
an object detector is first applied, and the remaining task is to 
associate 
the generated detections.
In this scheme, the most important issue is data association, which is 
mostly 
based on appearance similarity and motion consistency.
The motion features are mainly based on linear and nonlinear motion 
models. 
The linear model assumes the target to have a linear movement with 
constant 
velocity for a period of 
time~\cite{Dehghan2015GMMCP,xiang2015learning,Ristani2018Features},
which is used in most existing trackers. The nonlinear one, to some 
extent, 
can better capture the various movements and provide a more accurate 
motion 	prediction~\cite{yang2012multi-target,yang2012an}.
Many previous works on MOT try to develop more powerful appearance 
feature for object association, from the hand-crafted appearance features such 
as color histograms~\cite{Zamir2012GMCP,Dehghan2015GMMCP}, to the recent 
deep network based appearance features~\cite{chu2019famnet,xu2019spatial,xu2020train}.
This way, a key issue for such tracking-by-detection methods lies in the 
learning of human appearance features.
More recent works also try to achieve object detection and tracking 
simultaneously using an end-to-end 
{framework~\cite{zhou2020tracking,2019Tracking,wu2021track,zhang2022bytetrack}, e.g., making the detection and tracking to complement each other~\cite{wu2021track}, or excavating useful information from the low-confidence detection boxes~\cite{zhang2022bytetrack}}.
%{framework~\cite{zhou2020tracking,2019Tracking}}, or excavate useful information from the low-confidence detection boxes~\cite{zhang2022bytetrack}, or make the detection and tracking to complement each other~\cite{wu2021track}.
More recently, some trackers based on graph neural networks (GNN), e.g.,~\cite{braso2020learning} or transformers, e.g.,~\cite{meinhardt2022trackformer,ma2022unified,zhou2022global} are proposed and achieve the promising performance.
For a more comprehensive review on MOT, we refer readers to some 
excellent surveys on 
tracking~\cite{smeulders2013visual,ciaparrone2020deep}.
Note that, the problem studied in this paper is based on the MOT 
technique but 
not focus on the study of general MOT.

\textbf{MTMCT} (Multi-Target Multi-Camera Tracking) is an extension of 
MOT, 
which aims to track and re-identify the targets (mainly for humans) in a 
large field, e.g., a campus, using many cameras installed at many sites 
with 
little or no field of view overlap. 
Some related works in this 
area~\cite{gilbert2006tracking,prosser2008multi,cai2014exploring,chen2014object}
focus on the inter-camera tracklets association by assuming that the 
within-camera tracklets in each camera are prior given or obtained by 
existing algorithms. 
This setting is not practical in a real-world application.
Some other works aim to address a more realistic problem by solving both 
intra- and inter-camera tracking 
jointly~\cite{2016Performance,maksai2017non,Ristani2018Features,tesfaye2019multi}.
The main thought for solving such a problem is to learn more 
discriminative appearance features~\cite{Ristani2018Features} or design 
a 
more exquisite optimization model~\cite{tesfaye2019multi}.
Differently, we are more interested in handling multi-human association 
and tracking problem to comprehensively observe a crowd from different perspectives as discussed below. 

%	\textbf{MvMHAT} is different from MTMCT. 	
\textbf{MvMHAT} and MTMCT both stem from MOT task. 
However, they are different in that the former focuses on both the \textit{over-time human tracking} and the \textit{cross-view association at each time among multi-view videos}, where the multiple cameras pay co-attention on an overlapped area. The latter emphasizes the \textit{long-term} human tracking and re-identification (cross-camera association) using multiple cameras covering different areas. 
Two classic differences are 1) They have different problem definitions. Besides single-view intra-camera tracking, MTMCT also aims to handle the human re-identification, which is a ranking problem. Differently, MvMHAT focuses on the frame-by-frame multi-human cross-view association, which is a multi-graph matching problem. 
2) They use different camera settings. MTMCT uses multiple cameras distributing at different sites in a large-scale area with no field-of-view (FOV) overlap. Differently, MvMHAT uses multi-view all-around cameras with overlapping FOVs covering the same scene.	

Some early works~\cite{2008Multicamera,ayazoglu2011dynamic,khan2006multiview,leal2012branch,hofmann2013hypergraphs,eshel2010tracking} similar to MvMHAT
have studied the similar problems of tracking multiple humans using 
several overlapped cameras, in which the subjects commonly appear in different 
views at the same time.	
Recently, a series of works proposed by Xu et al.~\cite{xu2016multi-view,xu2017cross,liu2017stochastic} propose the 
tracking of multiple people in a scene, e.g., a garden, using several 
cameras and collect new datasets for research. This series of works develop 
various human features for tracking, including the varied poses and human 
actions, etc. 
However, the above works mainly focus on the over-time human tracking 
performance but not assessing the cross-view human association results.
A recent work~\cite{3Dposetracking} proposes to handle the multi-view human association and 3D pose estimation problem, in which a 3D pose tracking is integrated to obtain the temporal human pose.
Similarly, Dong et al.~\cite{Ass4Tracking} develop a self-supervised method for human association, which adapts the generic person appearance descriptor to the unlabeled videos by exploiting motion tracking, mutual exclusion constraints, and multi-view geometry.
These couple of works both require the camera calibration, which is not practicable in this work where the camera may keep moving over time.
More recently, a series of works by Han et al.~\cite{han2021multiple,Han2019Complementary} propose to jointly solve 
the human association and tracking problem using two complementary views, 
which, however, is a specific setting and used for several application 
scenarios. Differently, in this paper, we focus on a more general setting where 
(arbitrary) multiple cameras observe a scene from different views.

Also related to our work is a study on multi-view multi-object 
association 
(matching)~\cite{dong2019fast,arxiv2019Multiple,Zheng2016cross-view,han2020cvid,han2020cip}
by exploring the matching cues, including human 
appearance~\cite{han2020cvid,dong2019fast}, spatial 
relation~\cite{han2020cip,arxiv2019Multiple} or 
motions~\cite{Zheng2016cross-view}, all of which only focus on 
cross-view 
association but not involving the over-time tracking. 

%\textbf{Multi-view multi-human analysis.} detection, association, pose estimation and tracking, activity understanding

\begin{figure*}[ht]
%	\vspace{-0.3cm}
	\centering
	\includegraphics[width=1\linewidth]{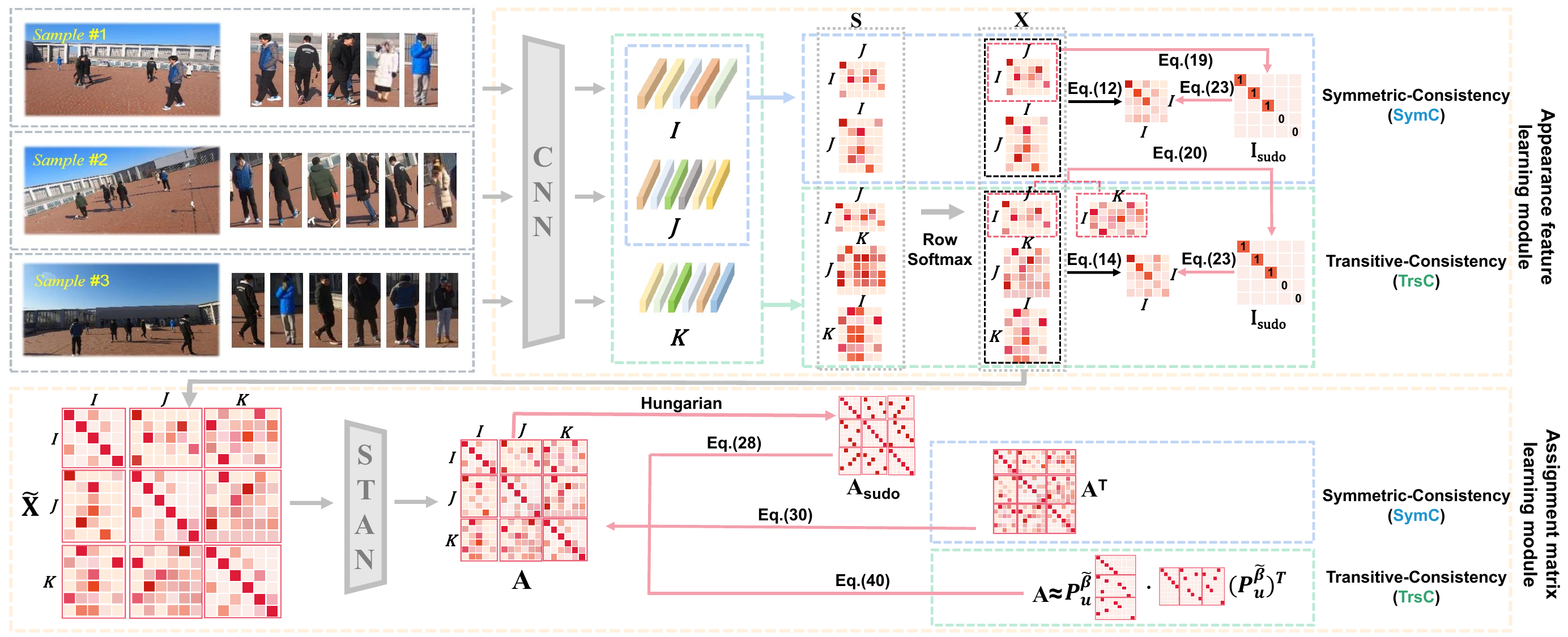}
%	\vspace{-0.3cm}
	\caption{Overall framework of the proposed method, {where we take three frames as an example. Specifically, {we use the frames from different views and time and their human detection results as an input batch in the training stage}. The framework consists of two parts: appearance feature learning module and assignment matrix learning module, for each of which we use the symmetric-consistency and transitive-consistency discussed in Section~\ref{sec:idea} to construct the self-supervised loss for training.}}
	\label{fig:2}
%	\vspace{-0.4cm}
\end{figure*}

\textbf{Self-supervised Learning} is a form of unsupervised learning, which has been widely used in many vision and multimedia 
computing tasks, including the image-based representation 
learning~\cite{zhang2016colorful,doersch2015unsupervised} and 
video-level 
temporal coherence~\cite{wang2019learning,lai2019self}, in particular, 
including person re-identification 
(re-id)~\cite{li2019unsupervised,li2018unsupervised,wu2019unsupervised} 
which is similar to our problem that learns the appearance similarity.
The global matching based re-id features can not be directly used for 
MOT, 
which is regarded as a local association problem~\cite{hou2019locality}.
However, there are rare works on studying the unsupervised learning 
based 
MOT, especially for the multi-human 
tracking~\cite{karthik2020simple,ho2020unsupervised}, not to mention the 
MvMHAT. 
Specifically, as one branch of unsupervised learning, self-supervised 
learning aims to construct the pretext tasks, commonly obtained by the 
acknowledged prior or self-constraint, to learn the network from 
unlabeled 
data.
In this paper, we aim to unveil the power of self-supervised learning 
for 
data association in MvMHAT.

\section{The Method}

\subsection{Problem Formulation}
\label{sec:Problem}

Given $V$ synchronized video sequences taken from different views, we 
aim to achieve the {multi-view multi-human association and tracking} (MvMHAT), which collaboratively tracks all subjects in all videos as well as identifies all the same person appearing in different views.
Specifically, we assume that the subjects have been detected in each 
frame in advance: the subjects are represented as bounding boxes in each view 
$v$ at each time $t$. 
For each person in each view, MvMHAT aims to connect the subjects to 
form 
the single-view trajectory. Besides, MvMHAT also identifies each 
trajectory belonging to the same person in all views.

In this work, we formulate the above collaborative tracking, i.e., 
MvMHAT, as a spatial-temporal subject association problem. 
On one hand, the temporal (over-time) association can be regarded as a 
single-view multiple object tracking (MOT) problem. 
Similar with most MOT approach, the goal is to solve the association 
matrix between the tracklets $\mathcal{T}_{t-1}$ {until} frame $t-1$ {in view $v$}, and all the detections $\mathcal{B}^{v}_{t} = \{B_i | i = 1,2,...,N_t^v\}$ 
on 
frame $t$. 
Thus the association matrix is represented as $\mathbf{A}_t^{v} \in 
\mathbb{R}^{M_{t-1} \times N_t^v}$,  where $M_{t-1}$ and $N_t^v$ denote 
the number of trajectories $\mathcal{T}_{t-1}$ and subjects 
$\mathcal{B}^{v}_{t}$, respectively.
On the other hand, the spatial (cross-view) association is a multi-view 
subject matching problem. 
At each time $t$, we establish the association relation between 
different 
views. Taking a pair of views $v$ and $u$ as an example, the cross-view 
subject association between $\mathcal{B}^{v}_{t}$ and 
$\mathcal{B}^{u}_{t}$ can be represented as a matching matrix 
$\mathbf{A}^{v,u} \in \mathbb{R}^{N_t^v \times N_t^u}$, where $N_t^u$ 
denotes the number of subjects in $\mathcal{B}^{u}_{t}$.

\subsection{Spatial-Temporal Self-Consistency}
\label{sec:idea}

\begin{figure}[ht]
	\centering
	\includegraphics[scale=0.25]{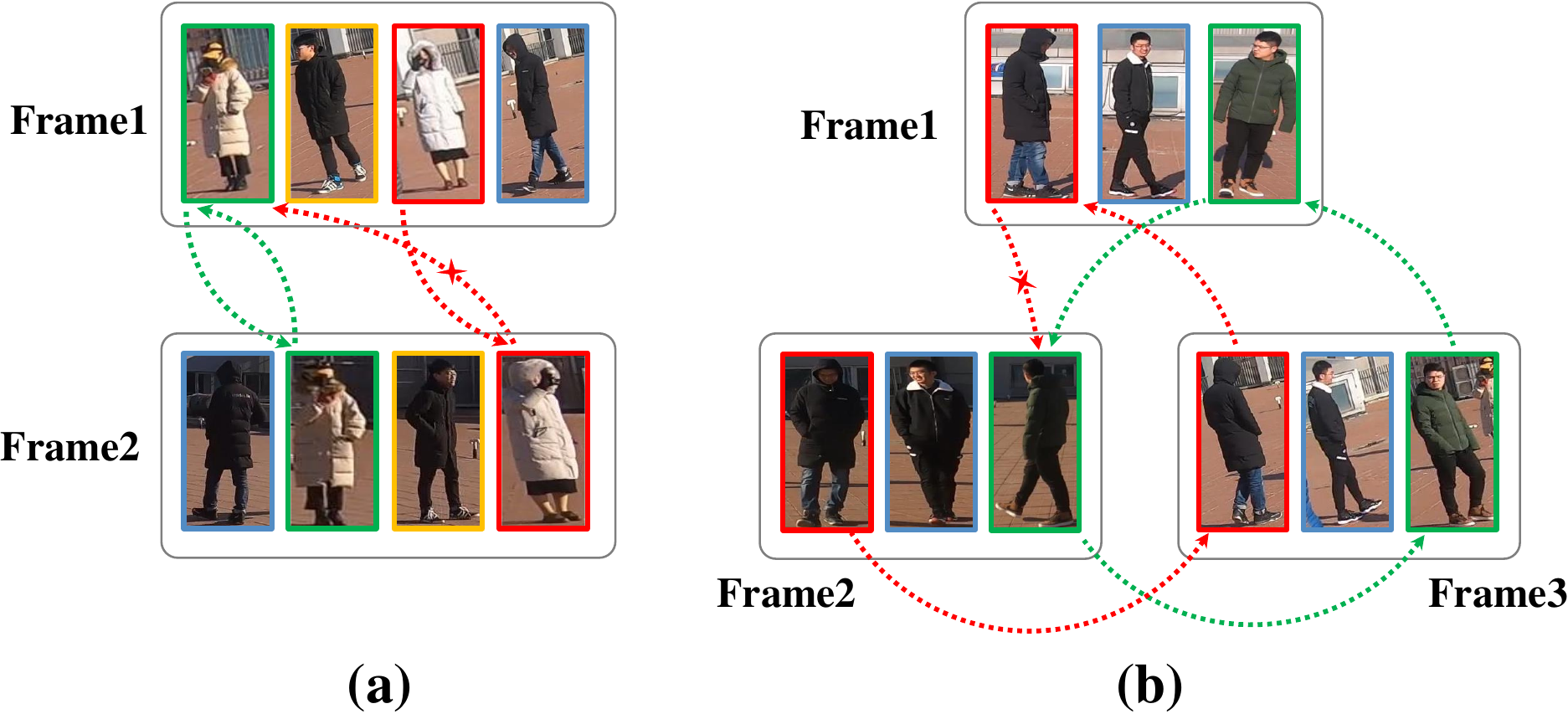}
	%	\vspace{-0.2cm}
	%\caption{{An illustration of the rationale of symmetric and transitive %self-consistency.}}
	\caption{{An illustration of symmetric and transitive consistency rationale.}}
	\label{fig:SIM}
	%	\vspace{-0.3cm}
\end{figure}

As shown in Fig.~\ref{fig:SIM}, the multi-view video sequences provide the all-around and time-varying appearance of the subjects in the scene. 
The same person appearing in pairwise views or points of time presents symmetric-consistency in Fig.~\ref{fig:SIM}(a). They also show transitive-consistency among 
multiple views and time in Fig.~\ref{fig:SIM}(b).
This inspires us to unveil self-supervised power for establishing the over-time and cross-view subject similarity.
In the following, we elaborate on these consistencies and develop a self-supervised learning network for spatial-temporal subject association.

The basic idea of the proposed self-supervised learning strategy is the Spatial-Temporal Self-Consistency contained in the MvMHAT problem.
We consider two pretext tasks to construct the self-supervised loss.
First, the same person appearing in a pair of frames from different 
views or different time should be consistent. 
As shown in Fig.~\ref{fig:SIM}(a), for a subject \#$A$ in Frame \#1, if the subject \#$A'$ in another Frame \#2 has the highest 
{similarity} with \#$A$ among all subjects; the subject \#$A$ should be the most {similar} with \#$A'$ among all subjects in Frame \#1, i.e., 
the symmetric-consistency property.
Second, as shown in Fig.~\ref{fig:SIM}(b), the subject has transitive 
consistency among multiple views and different time. 
For example, if subject \#$A$ in Frame \#1 is identity-consistent with \#$A'$ in Frame \#2 and the subject \#$A''$ in Frame \#3, then the subjects 
\#$A'$ and \#$A''$ should be the same subject,  i.e., the transitive-consistency property.

From the above observation, we obtain the following mathematical formalization: let $\mathcal{B} =\cup_{\substack{v=1...V\\t=1...T}} \mathcal{B}^{v}_{t}$, we define a subject
%
%\noindent
consistency relation $\mathcal{R} =\{(B,B')|B, B' \in \mathcal{B}, B$ and $B'$ denote the \textit{same} person$\} \subseteq \mathcal{B} \times \mathcal{B}$, {where $\times$ represents Cartesian product.} Clearly $R$ satisfies the following properties
\begin{align}
\label{eq:Reflexive}
	&\textbf{(1) {Reflexivity}: } (B,B) \in \mathcal{R}, \quad \forall B \in \mathcal{B} \\
	\label{eq:Symmetry}
	&\textbf{(2) Symmetry: } (B,B') \in \mathcal{R} \Rightarrow (B',B) \in \mathcal{R}, \quad \forall (B,B') \in \mathcal{R} \\
	\label{eq:Transitivity}
	&\textbf{(3) Transitivity: } (B,B') \in \mathcal{R}, (B',B'') \in \mathcal{R}  
	\Rightarrow  (B,B'') \in \mathcal{R}, 	\notag \\ &\qquad \qquad\qquad\qquad\qquad \qquad \qquad \forall (B,B'),(B',B'') \in \mathcal{R}
\end{align}

We simplify the subject set in view $v$ {at} time $t$, i.e., $\mathcal{B}^{v}_{t}\subseteq \mathcal{B}$, as $\mathcal{B}_i$.
We first assume that all $\mathcal{B}_i$ share the same set of subjects. 
We know this is difficult to hold in reality, and we will consider the more general case later.
We then denote the subject consistency relation between arbitrary set pair $(\mathcal{B}_i,\mathcal{B}_j)$ as $\varphi_{\mathcal{B}_i \mathcal{B}_j} \subseteq \mathcal{R}$, i.e., $\varphi_{\mathcal{B}_i \mathcal{B}_j} =\{(B,B')|B \in \mathcal{B}_i, B' \in \mathcal{B}_j, B$ and $B'$ denote the \textit{same} person$\}$. Then from Eqs.~(\ref{eq:Reflexive}-\ref{eq:Transitivity}) we can get
\begin{align}
	\label{eq:Reflexivemap}
	&\textbf{(1) {Reflexivity}: } &\varphi_{\mathcal{B}_i \mathcal{B}_i}=id_{\mathcal{B}_i}\\
	\label{eq:Symmetrymap}
	%\varphi_{\mathcal{B}_j \mathcal{B}_i} \circ \varphi_{\mathcal{B}_i \mathcal{B}_j}=id\\
	&\textbf{(2) Symmetry: } & \varphi_{\mathcal{B}_i \mathcal{B}_j} \triangleleft \varphi_{\mathcal{B}_j \mathcal{B}_i}=id_{\mathcal{B}_i}\\
	\label{eq:Transitivitymap}
	%\varphi_{\mathcal{B}_k \mathcal{B}_i} \triangleleft \varphi_{\mathcal{B}_j \mathcal{B}_k} %\triangleleft \varphi_{\mathcal{B}_i \mathcal{B}_j}=id
	&\textbf{(3) Transitivity: } & \varphi_{\mathcal{B}_i \mathcal{B}_j} \triangleleft \varphi_{\mathcal{B}_j \mathcal{B}_k} \triangleleft \varphi_{\mathcal{B}_k \mathcal{B}_i}=id_{\mathcal{B}_i}
\end{align}
where $\triangleleft$ denotes the left outer join operation to connect multiple $\varphi$, and $id_{\mathcal{B}_i}$ is the identity {mapping} of $\mathcal{B}_i$. 
This is to say the properties of {reflexivity}, symmetry and transitivity also maintain among the consistency relations $\varphi$.
In the following, we propose to use above self-consistency properties for subject feature learning in Section~\ref{sec:appfeature} and assignment matrix solving in Section~\ref{sec:assmatrix}.
%\redcolor{, that is the left relation is used as the basis to join the right relation, take $\varphi_{\mathcal{B}_i \mathcal{B}_j} \triangleleft \varphi_{\mathcal{B}_j \mathcal{B}_k} \subseteq \mathcal{B}_i \times \mathcal{B}_k$ as an example, all the $\mathcal{B}_i$ in the left relation appear in the result, with $\mathcal{B}_j$ equal as the join condition, if no element in the right relation meets the join condition, then $\mathcal{B}_k$ field is null} 

\subsection{Self-Consistency for Appearance Feature Learning}
\label{sec:appfeature}

%\textbf{Symmetrical-Consistency (SCON)}

%\begin{equation}
%\label{Eq:2}
%\mathbf{P} =
%\left( \begin{array}{cccc}
	%\mathbf{P}_{11} & \mathbf{P}_{12} & \ldots &  \mathbf{P}_{1C} \\
	%\mathbf{P}_{21} & \mathbf{P}_{22} & \ldots & \mathbf{P}_{2C} \\
	%\vdots & \vdots & \ddots & \vdots \\
	%\mathbf{P}_{C1} & \mathbf{P}_{C2} & \ldots & \mathbf{P}_{CC}
	%\end{array} \right),
	%\end{equation}
	\subsubsection{Feature Extraction}
	
	The overall framework is shown in Fig.~\ref{fig:2}, which takes the video sequence without annotation as input and learns the subject similarity used for association in a self-supervised manner.  
	Specifically, given an arbitrary video frame $i$, we first apply a human detector to 
	obtain 
	all the subjects $\mathcal{B}_i$ in this frame.
	With the detected subjects, we apply the feature extraction network, 
	denoted as $\Phi$, to get the feature representation for all subjects 
	$\mathbf{E}_i = \Phi (\mathcal{B}_i)$,
	by which we get $\mathbf{E}_i \in \mathbb{R}^{N_i \times D}$, here 
	$N_i$ denotes the number of subjects in frame $i$, and $D$ 
	denotes the dimension of feature for each subject. 
	%That is $\mathbf{E}_t^v$ is composed of each subject's feature 
	%${\mathbf{e}}_i$ as one row namely $ \mathbf{E}^v_t = [{\mathbf{e}_1; 
		%\mathbf{e}_2; ..., \mathbf{e}_{N_t^v}}] $.
	With the extracted features on each frame, we can then define the 
	subject 
	similarity and association across views or over time. 
	\\
	{
	\textbf{Spatial-temporal association.} Given a pair of frames $i,j$ from different views or time, the subject similarity matrix among all subjects in the respective frame can be calculated by
	\begin{equation}
		\label{eq:spatial-temporal}
		\mathbf{S}_{ij} = \mathbf{E}_i  \cdot 
		(\mathbf{E}_j)^\text{T}   \in \mathbb{R}^{N_i \times N_j},
	\end{equation}
	whose value at $r$-th row and $c$-th column, i.e., $\mathbf{S}_{i,j}(r,c)$ represents the similarity between $r$-th subject in 
	$\mathcal{B}_i$ and $c$-th subject in $\mathcal{B}_j$.
	}
%	\bluecolor{
%	\textbf{Spatial association.}
%	Given a pair of frames from the same point of time $t$ but different 
%	views 
%	$v$ and $u$, the subject similarity matrix among all subjects in the respective frame can be calculated by
%	\begin{equation}
%		\label{eq:s-spatial}
%		\mathbf{S}^{v, u}_{t} = \mathbf{E}^v_t  \cdot 
%		(\mathbf{E}^{u}_t)^\text{T}   \in \mathbb{R}^{N_t^v \times N_t^u},
%	\end{equation}
%	whose value at $i$-th row and $j$-th column, i.e., $\mathbf{S}^{v, 
%		u}_{t}(i,j)$ represents the similarity between $i$-th subject in 
%	$\mathcal{B}^v_t$ and $j$-th subject in $\mathcal{B}^u_t$.
%	\\
%	\textbf{Temporal association.} Similarly, given a pair of frames from 
%	the 
%	same view $v$ but different points of time $t$ and $s$, the 
%	subject similarity matrix can be calculated by
%	\begin{equation}
%		\label{eq:s-temporal}
%		\mathbf{S}^v_{t, s} = \mathbf{E}^v_t \cdot 
%		(\mathbf{E}^{v}_s)^\mathrm{T}    \in \mathbb{R}^{N_t^v \times N_s^v}.
%	\end{equation}
%}
	We then consider obtaining the pairwise matching matrix {$\mathbf{X}_{ij} \in [0,1]^{N_i \times N_j}$}
	based on the above similarity matrix.
	For clarity, we provisionally simplify the similarity and matching matrices $\mathbf{S}_{ij}$ and $\mathbf{X}_{ij}$ as $\mathbf{S}$ and $\mathbf{X}$.
	We use a temperature-adaptive \textit{softmax} operation 
	${f}$~\cite{hinton2015distilling} to compute the matching matrix as
	\begin{equation}
		\label{eq:softmax}
		\mathbf{X}(r,c) = {f}_{r,c}(\mathbf{S}) = \frac{\exp(\tau 
			\mathbf{S}(r,c))}{\sum_{c'=1}^C \exp(\tau \mathbf{S}(r,c'))},
	\end{equation}
	where $r,c$ denote the index of row and column in $\mathbf{S}$ and $C$ 
	is 
	the number of columns for $\mathbf{S}$. That is we apply the 
	\textit{softmax} operation on each row of the matrix $\mathbf{S}$ and 
	get 
	$\mathbf{X}$ with same size as $\mathbf{S}$ taking values in $[0,1]$, as 
	shown in Fig.~\ref{fig:2}. 
	In Eq.~(\ref{eq:softmax}), we use the \textit{softmax} with a adjustable value 
	$\tau$ 
	as the adaptive temperature 
	\begin{equation}
		\label{eq:tau}
		\tau = \frac{1}{\epsilon} \log [\frac{\delta(C-1) + 1}{1 - \delta}],
	\end{equation}
	to control the soften ability of the function, where $\epsilon$ and 
	$\delta$ are two {pre-set} parameters. 
	
	So far, we get the predicted matching matrix $\mathbf{X}$. If we have 
	the 
	annotated data with a human identification label, the network can be 
	trained with the supervision of ground-truth $\mathbf{X}$.
	In this paper, we aim to explore the {spatial-temporal self-consistency} for 
	learning the network without manual labels.
	
	\subsubsection{Self-Consistency Learning}
	\label{sec:loss}
	
	\textbf{Symmetric-Consistency (SymC).}
	Given the similarity matrix between the subjects within two sets 
	$\mathcal{I}$ and $\mathcal{J}$ (including the cross-view or over-time 
	cases) as defined in Eq.~(\ref{eq:spatial-temporal}), which we denoted as $\mathbf{S}_{ij} \in 
	\mathbb{R}^{|\mathcal{I}| \times |\mathcal{J}|}${, where $|\cdot|$ denotes the number of elements in the set}. We apply the 
	\textit{softmax} operation on the similarity matrix $\mathbf{S}$ to get 
	the	matching matrix defined in Eq.~(\ref{eq:softmax}) as
	\begin{equation}
		\label{eq:Xij}
		\mathbf{X}_{ij} = {f}({\mathbf{S}}_{ij}) \in 
		\mathbb{R}^{|\mathcal{I}| \times |\mathcal{J}|}.
	\end{equation}
	The matching matrix $\mathbf{X}_{ij}$ can be regarded as a mapping 
	(matching relation) from $\mathcal{I}$ to $\mathcal{J}$, i.e., 
	$\mathcal{X}_{ij} : \mathcal{I} \mapsto \mathcal{J}$.
	Specifically, the row sum in $\mathbf{X}$ is equal to 1, and we can find 
	the maximum in each row of $\mathbf{X}$ to seek the matched subject of 
	someone in $\mathcal{I}$ from $\mathcal{J}$.
	Similarly, we get the mapping from $\mathcal{J}$ to $\mathcal{I}$ as
	\begin{equation}
		\mathbf{X}_{ji} = {f}({\mathbf{S}}_{ij}^{\mathrm{T}}) \in 
		\mathbb{R}^{|\mathcal{J}| \times |\mathcal{I}|}.
	\end{equation}
	%where ${\mathbf{S}}_{ij}^{\mathrm{T}}$ denotes the transposition of 
	%${\mathbf{S}}_{ij}$.
	{
	According to the `symmetry' property in Eq.~(\ref{eq:Symmetrymap}), we calculate the 
	\textit{symmetric-similarity matrix} 
	}
	\begin{equation}
		\label{eq:Is}
		\textbf{I}_{\mathrm{S}} = \mathbf{X}_{ij} \cdot \mathbf{X}_{ji} \in 
		\mathbb{R}^{|\mathcal{I}| \times |\mathcal{I}|},
	\end{equation}
	where $\mathbf{I}_{\mathrm{S}}$ can be regarded as the mapping: 
	$\mathcal{I} \mapsto \mathcal{J} \mapsto \mathcal{I}$. Ideally, if the 
	subjects in $\mathcal{I}$ and $\mathcal{J}$ are same, the result 
	{$\mathbf{I}_\mathrm{ideal}$ (i.e., the ideal approximation matrix (ground-truth result) of $\mathbf{I}_{\mathrm{S}}$)} should be an identity matrix. 
	This way, we can calculate the loss between predicted 
	$\mathbf{I}_{\mathrm{S}}$ and the identity matrix.
	As discussed above, this assumption is not always satisfied due to the occluded or 
	out-of-view subjects over time and the field-of-view difference across 
	views, { which causes the all-zero row appearing in {$\mathbf{I}_\mathrm{ideal}$}.
	Therefore, we can not always compel the result $\mathbf{I}_{\mathrm{S}}$ 
	to approximate the identity matrix and have to apply more deliberate 
	supervision on it, which will be discussed later.
	
	%\begin{equation}
	%\mathbf{S}  = \mathbf{S}_{ij} ~~ (i \neq j).
	%\end{equation}
	
	\textbf{Transitive-Consistency (TrsC).} Besides the pairwise symmetric 
	similarity, we also consider the triplewise transitive similarity.
	%\bluecolor{Given the similarity matrix $\mathbf{S}_{ij}$ between two sets 
	%	$\mathcal{I}$ and $\mathcal{J}$, and $\mathbf{S}_{jk}$ between 
	%	$\mathcal{J}$ and $\mathcal{K}$, we also consider the consistent 
	%	similarity among this triplet, i.e., $\mathcal{I}$, $\mathcal{J}$ and 
	%	$\mathcal{K}$. }
	%\bluecolor{We first compute the third-order similarity matrix as
	%	\begin{equation}
	%		\mathring{\mathbf{S}}_{ik}  = \mathbf{S}_{ij} \cdot \mathbf{S}_{jk}, 
	%		~~ (i \neq j \neq k)
	%	\end{equation}
	%	where $\mathring{\mathbf{S}}_{ik} \in \mathbb{R}^{|\mathcal{I}| \times 
	%		|\mathcal{K}|}$  actually represents the similarity between the subjects 
	%	in $\mathcal{I}$ and $\mathcal{K}$, through the bridge of $\mathcal{J}$.
	%	When then compute the  matching matrix as
	%	\begin{equation}
	%		\begin{aligned}
	%			\mathring{\mathbf{X}}_{ik} = 
	%			{f}(\mathring{\mathbf{S}}_{ik})  \in 
	%			\mathbb{R}^{|\mathcal{I}| \times |\mathcal{K}|}, \quad
	%			\mathring{\mathbf{X}}_{ki} = 
	%			{f}(\mathring{\mathbf{S}}_{ik}^{\mathrm{T}}) \in 
	%			\mathbb{R}^{|\mathcal{K}| \times |\mathcal{I}|},
	%		\end{aligned}
	%	\end{equation}
	%	where $\mathring{\mathbf{S}}_{ik}^{\mathrm{T}}$ denotes the 
	%	transposition 
	%	of $\mathring{\mathbf{S}}_{ik}$. The result $\mathring{\mathbf{X}}_{ik}$ 
	%	represents the mapping $\mathcal{I} \mapsto 
	%	\mathcal{J} \mapsto \mathcal{K}$. In contrast, 
	%	$\mathring{\mathbf{X}}_{ki}$ represents the mapping along $\mathcal{K} 
	%	\mapsto \mathcal{J} \mapsto \mathcal{I}$.}\\
	{Given the similarity matrix $\mathbf{S}_{ij}$ between two sets 
		$\mathcal{I}$ and $\mathcal{J}$, $\mathbf{S}_{jk}$ between 
		$\mathcal{J}$ and $\mathcal{K}$, and $\mathbf{S}_{ki}$ between 
		$\mathcal{K}$ and $\mathcal{I}$, we also consider the transitive 
		similarity within this triplet, i.e., $\mathcal{I}$, $\mathcal{J}$ and 
		$\mathcal{K}$. We first compute their corresponding matching matrix as
		\begin{equation}
			\begin{aligned}
				&\mathbf{X}_{ij} = 
				{f}(\mathbf{S}_{ij})  \in 
				\mathbb{R}^{|\mathcal{I}| \times |\mathcal{J}|}, \quad
				\mathbf{X}_{jk} = 
				{f}(\mathbf{S}_{jk}) \in 
				\mathbb{R}^{|\mathcal{J}| \times |\mathcal{K}|}, \\
				&\quad \quad \qquad \mathbf{X}_{ki} = 
				{f}(\mathbf{S}_{ki}) \in 
				\mathbb{R}^{|\mathcal{K}| \times |\mathcal{I}|}, 
				~~ (i \neq j \neq k)
			\end{aligned}
		\end{equation}
		The result $\mathbf{X}_{ij}$, $\mathbf{X}_{jk}$ and $\mathbf{X}_{ki}$  represent the mapping $\mathcal{I} \mapsto \mathcal{J}$, $\mathcal{J} \mapsto \mathcal{K}$ and $\mathcal{K} \mapsto \mathcal{I}$ respectively.
	}
	Then, according to the `transitivity' property in Eq.~(\ref{eq:Transitivitymap}), we calculate the 
	\textit{transitive-similarity matrix}  as
	%\bluecolor{
	%	\begin{equation}
	%		\textbf{I}_{\mathrm{T}} = \mathring{\mathbf{X}}_{ik}  \cdot 
	%		\mathring{\mathbf{X}}_{ki} \in \mathbb{R}^{|\mathcal{I}| \times 
	%			|\mathcal{I}|},
	%	\end{equation}
	%}
		\begin{equation}
			\label{eq:It}
			\textbf{I}_{\mathrm{T}} = \mathbf{X}_{ij}  \cdot 
			\mathbf{X}_{jk}  \cdot  \mathbf{X}_{ki} \in \mathbb{R}^{|\mathcal{I}| \times 
				|\mathcal{I}|},
		\end{equation}
	where $\mathbf{I}_{\mathrm{T}}$ can be regarded as the mapping: 
	$\mathcal{I} \mapsto \mathcal{J} \mapsto \mathcal{K} \mapsto \mathcal{I}$. 
	Similar with the matrix $\mathbf{I}_{\mathrm{S}}$ defined in 
	Eq.~(\ref{eq:Is}), we need to apply an appropriate supervision on both 
	$\mathbf{I}_{\mathrm{S}}$ and $\mathbf{I}_{\mathrm{T}}$, which will be elaborated in detail in Section~\ref{sec:lossfeat}.
	
	%\textbf{Loss function design.}
	%$(\mathbf{S}^{\circ})^{\mathrm{T}} = (\mathbf{S}_{ij} \cdot 
	%\mathbf{S}_{jk})'$

	\textbf{{Generalized TrsC.}}
	Up to now, we have present the pairwise symmetric similarity, the 
	triplewise transitive similarity constraints.
	{
	Actually, the transitive-consistency should be satisfied in the more general cases where more views or time are involved.}
	In the following, we discuss the generalization of the proposed 
	self-consistency constraints.
	
%	\bluecolor{
%	We denote the universal set of the index, e.g., $i, j$ in 
%	Eq.~(\ref{eq:Xij}), for subject group on each frame as $\mathcal{F}$ 
%	namely $i, j \in \mathcal{F}$. 
%	In order to associate the subjects between each pair of (over-time or 
%	cross-view) frames, we denote the subjects association mapping between 
%	arbitrary pair $(i,j)$ as $\mathcal{M} = \{\varphi_{i, j} | \forall i, j 
%	\in \mathcal{F} \}$.
%	We first assume all the frames share the same set of subjects. Thus $\Phi_{i, 
%		j}$ 
%	is a bijection. \\
%	Given $\forall f_1,f_n \in  \mathcal{F}$,  the mapping $\varphi_{f_1 
%		f_n}$ from $f_1$ to $f_n$ can be decomposed as the combination of any 
%	number of mappings.
%	}
	{
	As discussed in Section~\ref{sec:idea}, we first assume all the frames share the same set of subjects. 
	This way, given any $n(n \geq 3)$ frames, the transitive-consistency among $n$ frames requires
	}
	%For each frame of image, given a collection of bounding boxes 
	%$\mathcal{B}$ and corresponding embedding feature vectors $\mathbf{E}$. 
	% Our goal is to update the map collection $\mathcal{M}$ to make 
	%association results better.
	%Given $\forall \varphi \in \mathcal{M}$, which can be represented as 
	%$\Phi_{f_1 f_n}$ with $f_1,f_n \in  \mathcal{F}$.
	%Actually, such mapping $\Phi$ can be decomposed as the combination of 
	%any number of mappings
	\begin{equation}
		\label{eq:phi}
		%\varphi_{\mathcal{B}_1 \mathcal{B}_n} = \varphi_{\mathcal{B}_{n-1} \mathcal{B}_n} \circ %\varphi_{\mathcal{B}_{n-2} 
		%	\mathcal{B}_{n-1}} \circ \cdots \circ \varphi_{\mathcal{B}_1 \mathcal{B}_2}, 
		\varphi_{\mathcal{B}_1 \mathcal{B}_n} = \varphi_{\mathcal{B}_1 \mathcal{B}_2} \triangleleft \cdots \triangleleft \varphi_{\mathcal{B}_{n-2} 
			\mathcal{B}_{n-1}} \triangleleft \varphi_{\mathcal{B}_{n-1} \mathcal{B}_n}.
	\end{equation}
%	\bluecolor{
%	where $f_1, f_2, ..., f_n \in \mathcal{F}$, and $\circ$ denotes the 
%	mapping composition operation.}
	The above combined mapping can be derived by the 
	proposed pairwise and triplewise mapping.\\
	\textit{\textbf{{Inference.}}} Equation~(\ref{eq:phi}) {can be guaranteed by} 
	\begin{equation}\label{eq:phi|ijk}
		\varphi_{\mathcal{B}_i \mathcal{B}_k} = \varphi_{\mathcal{B}_i \mathcal{B}_j} \triangleleft \varphi_{\mathcal{B}_j \mathcal{B}_k}, ~~ \mathrm{for}  
		~~\forall \mathcal{B}_i,\mathcal{B}_j,\mathcal{B}_k \subseteq \mathcal{B},
	\end{equation}
	from which we discuss the situations in three cases:\\
	{
	(i) $i = k = j$: it is equivalent to the {Reflexivity} in Eq.~(\ref{eq:Reflexivemap}).\\
	(ii) $i = k \neq j$: it is equivalent to the Symmetry in Eq.~(\ref{eq:Symmetrymap}).\\
	(iii) $i \neq k \neq j$: it is equivalent to the Transitivity in Eq.~(\ref{eq:Transitivitymap}).\\
	}Note that, these three cases cover all the scenarios.
	The {reflexivity} condition naturally holds in our problem because the 
	same person in the same frame has the same feature vector. 
	This way, we only consider the conditions in (ii) and (iii), for which we use the doubly stochastic matching matrix 
	$\mathbf{X}_{ij}$ to represent the consistency relation $\varphi$. We get
	\begin{align}
		%\begin{equation}
		&\text{(Symmetric-Similarity)} &\mathbf{X}_{ij} \cdot 
		\mathbf{X}_{ji} 
		= \mathbf{I}\\
		%\end{equation}
		%\begin{equation}
		&\text{(Transitive-Similarity)} & \mathbf{X}_{ij} \cdot 
		\mathbf{X}_{jk} \cdot \mathbf{X}_{ki} = \mathbf{I}
		%\end{equation}
	\end{align}
	where $\mathbf{I}$ is the identity matrix.  
	This way, the proposed method with self-supervision from symmetric-similarity {Eq.~(\ref{eq:Is})} and transitive-similarity {Eq.~(\ref{eq:It})}, can be regarded as the 
	self-consistency constraints for the matching relations among arbitrary $n$ frames.  $\hfill\blacksquare$ 
	{		
		
	In the above inference, we assume all the frames share the same set of subjects.
	Actually, this is not always satisfied in our problem.}
	Therefore, we only assume $\mathbf{X}$ is the row stochastic matrix (sum of 
	each row is 1). We also consider the dummy nodes and a relaxing strategy to construct the loss function for the supervision of $\mathbf{I}$, which is presented as below.
	
%	\bluecolor{use the margin in Eq.~(\ref{eq:margin}) to relax the 
%	identity matrix $\mathbf{I}$} \redcolor{use pseudo matrix label with dummy node to self-supervise $\mathbf{I}$}.

	%Actually, the training samples in our experiments cover any involved 
	%frame, i.e., $\forall i \in \mathcal{F}$.
	
	%\begin{equation}
	%f_{ii} = id
	%\end{equation}
	%
	%\begin{equation}
	%f_{ji} \circ f_{ij} = id
	%\end{equation}
	%
	%\begin{equation}
	%f_{ki} \circ f_{jk} \circ f_{ij} = id
	%\end{equation}

	%So the next step to think about is how do we determine
	%the diagonal elements value in a self-supervised manner. \\
	%
	%\redcolor{
		\subsubsection{Self-supervised Loss for Feature Consistency}
		\label{sec:lossfeat}
		Next, we uniformly denote the matrix $\mathbf{I}_{\mathrm{S}}$ in Eq.~(\ref{eq:Is}) and $\mathbf{I}_{\mathrm{T}}$ in Eq.~(\ref{eq:It}) as $\mathbf{I}$ provisionally. Its ideal approximation matrix $\mathbf{I}_\mathrm{ideal}$ have the property that their diagonal elements are 1 or 
		0, while other elements are all 0. 
		Note that, the element of 0 in the diagonal denotes that the subject is missing (due to the occlusion or out of view, etc) in one view.
		However, for the self-supervised method, we can not obtain the ideal approximation matrix $\mathbf{I}_\mathrm{ideal}$ for supervision.
		A simple method is to use a diagonal identity matrix to supervise $\mathbf{I}$ with a relaxation~\cite{gan2021self}, which ignores the case of missing subjects.
		Differently, in this work, we handle this problem by constructing a dummy nodes integrated pseudo matrix label to self-supervise $\mathbf{I}$.
		
		\textbf{Pseudo matrix label with dummy nodes.}
		First, let's think about the symmetric-consistency. 
		Given two sets $\mathcal{I}$ and $\mathcal{J}$, if a certain person in set $\mathcal{I}$ also appear in set $\mathcal{J}$, then after the mapping from $\mathcal{I} \mapsto \mathcal{J} \mapsto \mathcal{I}$, the diagonal element value of the corresponding row of this person in $\mathbf{I}_\mathrm{ideal}$ matrix should be 1; otherwise 0. 
		This way, we construct the pseudo matrix label as
		\begin{equation}
			\label{eq:pseudo1}
			\begin{split}		
				\mathbf{I}^\mathrm{S}_\mathrm{sudo} (r,r)= \left \{		
				\begin{array}{ll}			
					1,                    & \max\limits_{c} \mathbf{X}_{ij}(r,c) \textgreater {M} \\					
					0,                    & \mathrm{otherwise}			
				\end{array}
				\right.		
			\end{split}
		\end{equation}
		where $\mathbf{X}_{ij}$ is obtained by the row softmax operation on $\mathbf{S}_{ij}$ matrix as defined above. For each row in $\mathbf{X}_{ij}$ (i.e. each person in set $\mathcal{I}$) represents the matching scores with each person in set $\mathcal{J}$.
		Therefore, for each row in {$\mathbf{X}_{ij}$}, if the maximum value of the row is greater than the matching threshold ${M}$, it is considered that the person also appears in the set $\mathcal{J}$, and we can conclude that the diagonal {element} value of this row in $\mathbf{I}_\mathrm{sudo}$ is 1; otherwise 0. 
		
		%The above formulation can be implemented by using the $\mathbf{X}_{ij}$ matrix, which we call the judgement matrix here.  
		%Note that when the $\mathbf{X}_{ij}$ is reliable (i.e. $\mathbf{X}_{ij}$ can correctly reflect the matching relationship between the person in set $\mathcal{I}$ and set $\mathcal{J}$), this method is a sufficient and necessary condition for the judgement of the diagonal elements of the $\mathbf{I}_{ideal}$. 
		For the the transitive-consistency, given three sets $\mathcal{I}$, $\mathcal{J}$ and $\mathcal{K}$, if a certain person in set $\mathcal{I}$ also appear in set $\mathcal{J}$ and $\mathcal{K}$, the diagonal element value of the corresponding row for this person in $\mathbf{I}_\mathrm{ideal}$ matrix should be 1; otherwise 0.
		Similarly with Eq.~\eqref{eq:pseudo1}, we construct the pseudo matrix label as
		\begin{equation}
			\begin{split}		
				\mathbf{I}_\mathrm{sudo}^\mathrm{T}(r,r)= \left \{		
				\begin{array}{ll}			
					1,  &\max\limits_{c_{1}} \mathbf{X}_{ij}(r,c_{1}) \textgreater {M}  \wedge 
					\max\limits_{c_{2}} \mathbf{X}_{ik}(r,c_{2}) \textgreater {M}\\					
					0,  &\mathrm{otherwise}			
				\end{array}
				\right.		
			\end{split}
		\end{equation}
		where we use  $\mathbf{X}_{ij}$ and $\mathbf{X}_{ik}$ for judgement.
		
		%\redcolor{
			%Next let's think about the transitive-consistency. 
			%Similar to the idea of symmetric-consistency, Here we use. 
			%In fact, it can also use the mapping $\mathcal{I} \mapsto \mathcal{J} \mapsto \mathcal{K}$ result as the only judgement condition, but here we do not use the result of $\mathbf{X}_{ij}$*$\mathbf{X}_{jk}$ as the judgement matrix, because during the calculation, each row of $\mathbf{X}_{ij}$ is multiplied by each column in $\mathbf{X}_{jk}$, However, the $\mathbf{X}_{jk}$ matrix is obtained by row softmax, and its columns may not correctly reflect the matching relationship between the person in set $\mathcal{K}$ and each person in set $\mathcal{J}$. The diagonal elements of the $\mathbf{I}_{ideal}$ can be judged as below:
			%Here we don't explicitly require that the same person in set $\mathcal{J}$ as the certain person in set $\mathcal{I}$ and the same person in set $\mathcal{K}$ as that person in set $\mathcal{I}$ must also be the same person in the mapping from $\mathcal{J} \mapsto \mathcal{K}$. This is because when the $\mathbf{X}_{ij}$ and $\mathbf{X}_{ik}$ matrices are reliable (i.e. they can correctly reflect the matching relationship between people in sets $\mathcal{I}$ and $\mathcal{J}$, and sets $\mathcal{I}$ and $\mathcal{K}$, respectively), it is obvious that the above implicit conditions hold. Therefore, under the above conditions, this method is also a sufficient and necessary condition for the judgement of the diagonal elements of the $\mathbf{I}_{ideal}$.\\

			\textbf{Loss function design.} We also uniformly denote the pseudo matrices $\mathbf{I}_\mathrm{sudo}^\mathrm{S}$ and $\mathbf{I}_\mathrm{sudo}^\mathrm{T}$ as 
			$	\mathbf{I}_\mathrm{sudo} $.
			Here we do not directly use $\mathbf{I}_\mathrm{sudo}$ matrix to compute the difference with $\mathbf{I}$ as loss. This is because $\mathbf{X}_{ij}$  matrix is obtained by row softmax, it is impossible to have a row with all 0 values, as discussed in~\cite{gan2021self,wang2020cycas}. So, the off-diagonal elements values in $\mathbf{I}$ matrix, obtained by multiplication of $\mathbf{X}_{ij}$, are also not 0, 
			while the off-diagonal elements values in $\mathbf{I}_\mathrm{sudo}$ are all 0.
			The direct difference between them as loss might bring in additional error.
			This way, we design the loss function according to the characteristics of $\mathbf{I}_\mathrm{sudo}$.\\
			1) When the diagonal element of a row in $\mathbf{I}_\mathrm{sudo}$ is 1. We can easily find that its characteristic is that the value of the diagonal element is greater than its off-diagonal elements. So the diagonal element of $\mathbf{I}$ should be greater than the others in this row. With this constraint, we apply the loss function as below
			\begin{equation}
				\label{eq:margin}
				\mathrm{L}_1(\textbf{I}_r) = 
				\mathrm{relu}(\max_{c \neq r} \textbf{I}(r,c) - \textbf{I}(r,r) + m_{1})
			\end{equation}
			where $r, c$ denote the indices of row and column in $\mathbf{I}$}, i.e., {$\textbf{I}_r$ denotes the $r$-th row of matrix $\textbf{I}$.
			Specifically, for each row $r$, if the off-diagonal elements 
			$\textbf{I}(r,c)$ with $c \neq r$ are greater than the corresponding 
			diagonal element $\textbf{I}(r,r)$, the loss will 
			increase~\cite{wang2020cycas}.
			We only use the maximum off-diagonal element $\max_{c \neq r} 
			\textbf{I}(r,c)$ in each row to punish the hardest negative sample.
			The margin $m_{1} \geq 0$ is a pre-set parameter, which controls the 
			punishment scope for the gap between $\textbf{I}(r,c)$ and  
			$\textbf{I}(r,r)$. 
			In other word, the loss will take effect iff $\textbf{I}(r,r) - \max_{c 
				\neq r} \textbf{I}(r,c) \leq m_{1}$. This setting expects the diagonal 
			element to be greater than the other elements with a margin $m_{1}$. \\
			{
				2) When the diagonal element of a row in $\mathbf{I}_\mathrm{sudo}$ is 0. We can find that its characteristic is that the value of the diagonal element is close to the values of the off-diagonal elements. Following the above idea, we define the following loss
			}
			\begin{equation}	\label{eq:margin2}
				\begin{aligned}
					\mathrm{L}_2(\textbf{I}_r) = \frac{1}{2}(
					& \mathrm{relu}(\max_{c \neq r} \textbf{I}(r,c) - \textbf{I}(r,r) - m_{2})+ \\ 
					& \mathrm{relu}(\textbf{I}(r,r) - \min_{c \neq r}\textbf{I}(r,c) - m_{2})).
				\end{aligned}
			\end{equation}
			The margin $m_{2} \geq 0$ is also a pre-set parameter, which controls the punishment scope for the gap between $\textbf{I}(r,c)$ and $\textbf{I}(r,r)$. Here we want to have $\left| I(r,c)-I(r,r) \right| \leq m_{2},  \forall c \neq r$. When this condition isn't satisfied, we penalize {both} the upper ($\max$) and lower ($\min$) bounds of it.
				Therefore the loss function of $\textbf{I}$ is defined by combining the above two cases as
				\begin{equation}	\label{eq:l1l2}
					\begin{aligned}
						\mathrm{L}(\textbf{I}) = 
						\sum_{r=1}^{\left| \mathcal{I} \right|} \mathbf{I}_\mathrm{sudo}(r,r) \cdot \mathrm{L}_1(\textbf{I}_r) +
						(1-\mathbf{I}_\mathrm{sudo}(r,r)) \cdot \mathrm{L}_2(\textbf{I}_r).
					\end{aligned}
				\end{equation}

			%If $m = 1$, the loss compel the diagonal element 
			%If $m = 0 $, the loss  for any case where $\max \textbf{I}(r,l) > 
			%\textbf{I}(r,r)$, else 
			Finally, we define the SymC loss 
			\begin{equation}
				\label{eq:loss-s}
				\mathcal{L}_{\mathrm{Sym}}^\mathrm{A} =  \mathrm{L}(\textbf{I}_{\mathrm{S}} ) 
				+ 
				\mathrm{L}(\textbf{I}_{\mathrm{S}}^\text{T}),
			\end{equation}
			where $\mathrm{L}(\textbf{I}_{\mathrm{S}} )$ and 
			$\mathrm{L}(\textbf{I}_{\mathrm{S}}^\text{T})$ compels 
			$\textbf{I}_{\mathrm{S}}$ to satisfy above constraints for all rows and 
			columns, respectively. Similarly, we get the TrsC 
			loss
			\begin{equation}
				\label{eq:loss-t}
				\mathcal{L}_{\mathrm{Trs}}^\mathrm{A} =  \mathrm{L}(\textbf{I}_{\mathrm{T}} ) 
				+ 
				\mathrm{L}(\textbf{I}_{\mathrm{T}}^\text{T}).
			\end{equation}

			%\textbf{Self-Similarity (SSIM):}
			%
			%\begin{equation}
			%\mathbf{X}^{'}  = \mathbf{X}_{ij} (i \neq j)
			%\end{equation}
			%
			%\begin{equation}
			%\textbf{I}' = \mathbf{X}' \cdot (\mathbf{X}')^{\mathrm{T}}
			%\end{equation}
			%\\
			%\textbf{Transitive-Similarity (TSIM):}
			%
			%
			%\begin{equation}
			%\mathbf{X}^{''}  = \mathbf{X}_{ij} \cdot \mathbf{X}_{jk} (i \neq j \neq 
			%k)
			%\end{equation}
			%
			%$A \rightarrow B$  $B \rightarrow C$   $A \rightarrow B \rightarrow C$
			%
			%\begin{equation}
			%\textbf{I}'' = \mathbf{X}' \cdot (\mathbf{X}'')^{\mathrm{T}}
			%\end{equation}
			
\subsection{Self-Consistency for Assignment Matrix Learning}
\label{sec:assmatrix}
{
	In the above section, we have leveraged the self-consistency to construct the self-supervised loss for feature extraction.
	To handle the MvMHAT task, we aim to obtain the spatial-temporal subject assignment matrix involving multiple time and views.
	This way, in this section, we aim to further explore the global structure information in solving the assignment matrix, for which we also leverage the self-consistency properties as discussed in Section~\ref{sec:idea}, and design the corresponding constraints as self-supervised losses in our framework.	
%	aim to dig more  through combining the spatial-temporal matching matrices together, and get the 
%	The X matrix trained through the above self-supervised loss can be used as spatial association and temporal association assignment matrix after post-processing to obtain a binary matrix, which has already considered self-consistency. But it is not enough, we c
}
\subsubsection{Spatial-Temporal Global Assignment}
\label{sec:STGA}
{
	\textbf{Spatial-temporal affinity matrix generation.}
	From the above section, we can get the matching matrix $\mathbf{X}_{ij}$, which is obtained by the softmax operation through the feature similarity between the subjects in a pair of frames $i,j$. 
	In the inference stage, we can use $\mathbf{X}_{ij}$ to get the subject association results directly. However, it does not consider the information of the subjects in multiple views and multiple time. 
	For MvMHAT task, it is not a pairwise association problem, but a spatial-temporal jointly optimized subject association problem. 
	So the global information among multiple views and time is important and can help to get more reliable association. 
%	Like the general linear assignment problem, it requires an affinity matrix as input, and the output is an assignment matrix.
	%Considering the space occupied by the input matrix and the running speed of the assignment %algorithm, here we only use two consecutive time points and all views subjects to generate the %input matrix. And concatenation in the following order:\\
	This way, we consider to use the matching matrix from multiple time and all views to generate the a global assignment matrix. 
	Here we take arbitrary two points of time, i.e., time $t$ and $s$, and all $V$ views, and concatenate the subjects from them in the following order
	\begin{equation}
		\label{Eq:concate}
		\tilde{\mathcal{B}} =\mathcal{B}_{t}^1 \cup ... \cup \mathcal{B}_{t}^V \cup \mathcal{B}_{s}^1 
		\cup ... \cup \mathcal{B}_{s}^V.
	\end{equation}}

	Like before, we use feature extraction network to extract all subject feature in $\tilde{\mathcal{B}}$ and get $\tilde{\mathbf{E}}=\Phi(\tilde{\mathcal{B}})$. We then calculate similarity matrix $\tilde{\mathbf{S}} = \tilde{\mathbf{E}}  \cdot \tilde{\mathbf{E}}^\text{T}$. 
	We first split the rows and columns of the $\tilde{\mathbf{S}}$ according to the time and view which the subject belongs to, and then use Eq.~(\ref{eq:softmax}) to calculate the matching matrix.
	This way, we can get the spatial-temporal affinity matrix $\tilde{\mathbf{X}}$ as
%	each sub-matrix to obtain the final input matrix, i.e. . We can easily find that $\mathbf{A}$ matrix can also be concatenated through the matching matrix $\mathbf{X}$  calculated above
	\begin{equation}
		\label{Eq:combineX}
		\tilde{\mathbf{X}} =
		\left( \begin{array}{cccccc}
			\mathbf{X}_{t,t}^{1,1}&\ldots&\mathbf{X}_{t,t}^{1,V}&\mathbf{X}_{t,s}^{1,1}&\ldots& \mathbf{X}_{t,s}^{1,V} \\
			\vdots & \ddots & \vdots & \vdots & \ddots & \vdots \\
			\mathbf{X}_{t,t}^{V,1}&\ldots&\mathbf{X}_{t,t}^{V,V}&\mathbf{X}_{t,s}^{V,1}&\ldots& \mathbf{X}_{t,s}^{V,V} \\
			\mathbf{X}_{s,t}^{1,1}&\ldots&\mathbf{X}_{s,t}^{1,V}&\mathbf{X}_{s,s}^{1,1}&\ldots& \mathbf{X}_{s,s}^{1,V} \\
			\vdots & \ddots & \vdots & \vdots & \ddots & \vdots \\
			\mathbf{X}_{s,t}^{V,1}&\ldots&\mathbf{X}_{s,t}^{V,V}&\mathbf{X}_{s,s}^{V,1}&\ldots& \mathbf{X}_{s,s}^{V,V} \\
		\end{array} \right) \in \mathbb{R}^{|\tilde{\mathcal{B}}| \times |\tilde{\mathcal{B}}|},
	\end{equation}
	where $\mathbf{X}_{t,s}^{u,v}$ represents the matching matrix generated by the two sets $\mathcal{B}_{t}^u$ and $\mathcal{B}_{s}^v$.

\begin{figure}[t]	\vspace{-0cm}
	\centering
	\includegraphics[scale=0.203]{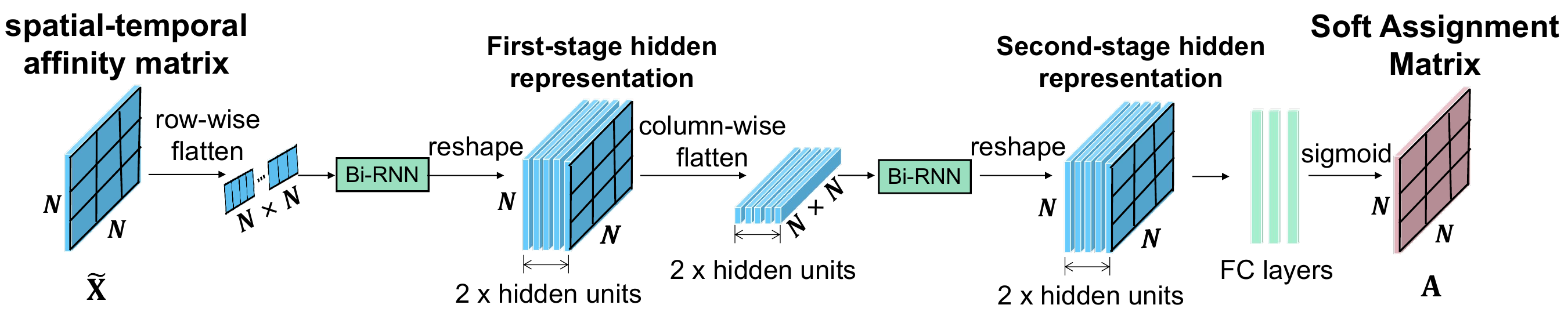}
	\caption{An illustration of the structure of STAN.}
	\label{fig:STAN_pic} 
	%	\vspace{-15pt}
\end{figure}

{
	\textbf{Spatial-temporal assignment network.}
	%Some existing assignment algorithms can't be directly used here to solve this problem, because they are used for pairwise matching, and the output is a doubly stochastic matrix. 
	To build an end-to-end framework to achieve assignment optimization from the input $\tilde{\mathbf{X}}$, we develop a deep neural network namely STAN (spatial-temporal assignment network) to get the final assignment matrix considering the global structure. 
	We require STAN to meet two conditions: 1) The network should be able to handle the inputs of different sizes, since the dimension of the input matrix $\tilde{\mathbf{X}}$ is determined by $\tilde{\mathcal{B}}$, whose size is always changing; 
	2) The network must have a global receptive field for the input matrix, since the output assignment matrix should satisfy the global optimization including over-time and cross-view assignment. 
}

Inspired by previous work~\cite{xu2020train}, we use bi-directional RNN network architecture to develop the STAN, which takes the affinity matrix $\tilde{\mathbf{X}}$ calculated above as input.
As shown in Fig.~\ref{fig:STAN_pic}, the input affinity matrix $\tilde{\mathbf{X}}$ is first reshaped as a 1D vector according to the row-wise order {flatten} and fed into a BiRNN. 
Next, the output of the BiRNN is then reshaped as the {2D matrix} according to the column-wise order {flatten} and fed into another BiRNN.
Later, we apply the network composed of three FC (fully-connected) layers and a Sigmoid function to obtain the final assignment matrix $\mathbf{A}$.
%At the same time, since the assignment matrix is a binary matrix, it is not differentiable, so we relax the output of the network into a real matrix with each element value between 0 and 1, we can use the sigmoid function to guarantee that.

\subsubsection{Self-supervised Loss}
\label{sec:selfsup}
We aim to design the self-supervised loss function to train STAN.
Specifically, the self-supervised loss is constructed from two aspects: 1) automatically generates pseudo label from each pair of frames that provides the basic assignment results, 2) auxiliary loss to constrain the network to include the global spatial-temporal structural information. 

%We hope that the , then we can use 
%Like supervised training, we need to provide a label to the  to provide the correct direction for network training. Since we have no ground truth labels available, we need to construct pseudo-labels. 

%\redcolor{
	\textbf{Pseudo label loss.}
	We first consider the pseudo label by processing each sub-matrix $\mathbf{X}_{t,s}^{u,v}$ in Eq.~(\ref{Eq:combineX}) individually to obtain a binary pseudo-label matrix. Specifically, for each $\mathbf{X}_{t,s}^{u,v}$, we use the Hungarian algorithm {with matching threshold $M$} to get the binary matrix as the one-to-one assignment result. We concatenate the corresponding binary matrix in the order of $\tilde{\mathbf{X}}$ to get the pseudo label ${\mathbf{A}_\mathrm{sudo}}$. Considering the positive and negative samples in ${\mathbf{A}_\mathrm{sudo}}$ are unbalanced, we use the focal loss~\cite{lin2017focal}
	\begin{equation}
		\label{Eq:Focalloss}
		\begin{aligned}
			\mathcal{L}_\mathrm{Pse}^\mathrm{M} =
			\left\{ \begin{array}{ll}
				-\alpha{(1-a_{rc})}^{\gamma}\log (a_{rc}) & , \textrm{if $\bar{a}_{rc}=1$}\\
				-(1-\alpha)(a_{rc})^\gamma \log(1-a_{rc}) & , \textrm{if $\bar{a}_{rc}=0$}
			\end{array} \right.
		\end{aligned}
	\end{equation}
	%where $a_{rc}$ and  $\bar{a}_{rc}$ represent the elements values of position $(r,c)$ in $\mathbf{A}$ and ${\mathbf{A}_\mathrm{sudo}}$, respectively. $\alpha$ and $\gamma$ are two pre-defined parameters.
	where $a_{rc}$,  $\bar{a}_{rc}$ represent the elements values of position $(r,c)$ in $\mathbf{A}$ and ${\mathbf{A}_\mathrm{sudo}}$, respectively. $\alpha$, $\gamma$ are two pre-defined parameters.

	\textbf{Symmetric-Consistency (SymC) loss.}
	Note that, the above pseudo label is obtained from each pairwise matching without considering any global structural constraint of the multiple time and views, which is considered in the following.
	We borrow the basic idea of symmetric-consistency in Eq.~(\ref{eq:Symmetry}) in Section~\ref{sec:idea}. Given sets $\mathcal{I}$ and $\mathcal{J}$, if a person in $\mathcal{I}$ has the highest similarity with someone in $\mathcal{J}$, then this person in $\mathcal{J}$ should also be the most similar to that person in $\mathcal{I}$. Then we have
	\begin{equation}
		\label{Eq:symloss}
		\begin{aligned}
			\mathbf{A}_{ij} & =\mathbf{A}_{ji}^\mathrm{T}, & \forall \mathcal{B}_i,\mathcal{B}_j \subseteq \tilde{\mathcal{B}},
		\end{aligned}
	\end{equation}
	where $\mathbf{A}_{ij}$ and $\mathbf{A}_{ji}$ sub-matrices are in symmetrical positions in $\mathbf{A}$, so Eq.~(\ref{Eq:symloss}) is equivalent to $\mathbf{A}=\mathbf{A}^\mathrm{T}$. Then we define the SymC loss for assignment matrix learning as
	\begin{equation}
		\label{Eq:SymC}
		\mathcal{L}_\mathrm{Sym}^\mathrm{M} =\|\mathbf{A}-\mathbf{A}^\mathrm{T}\|_2.
	\end{equation}
%}
\textbf{\textit{Discussion.}} The above constraint in Eq.~\eqref{Eq:symloss} has not been guaranteed by the pseudo label loss in Eq.~\eqref{Eq:Focalloss}. 
This is because $\mathbf{X}_{ij}$ matrix is obtained by row softmax from $\mathbf{S}_{ij}$, while the $\mathbf{X}_{ji}$ matrix is obtained by column softmax from $\mathbf{S}_{ij}$ with a matrix transpose, so $\mathbf{X}_{ij}$ is not equal to $\mathbf{X}_{ji}^\mathrm{T}$. Thus through the Hungarian algorithm, ${\mathbf{A}}_{ij}$ is not necessarily equal to ${\mathbf{A}}_{ji}^\mathrm{T}$. Therefore, we add the SymC loss in Eq.~\eqref{Eq:SymC} as the constraint.

\textbf{Transitive-Consistency (TrsC) loss.}
Inspired by the idea of transitive-consistency in Section~\ref{sec:idea}, given multiple sets $n$ ($n \geq 3$), the same person appearing in these sets should be associated as a closed loop. 
In this section, we design the transitive-consistency loss to constrain the assignment matrix to satisfy this condition.

For this purpose, we introduce a set $\mathcal{U}$, which contains all the subjects {with different identities} in $\tilde{\mathcal{B}}$ (i.e., the subjects from different frames but representing the same person denote the same element in $\mathcal{U}$).
Then we construct a {row-concatenation} permutation matrix (i.e., the binary assignment matrix between the elements in two sets%\greencolor{i.e., a matrix storing all mappings(matching relations)}
) from $\tilde{\mathcal{B}}$ to $\mathcal{U}$, i.e., $\mathbf{P}^{\tilde{\mathcal{B}}}_{\mathcal{U}} \in \mathbb{R}^{|\tilde{\mathcal{B}}| \times |{\mathcal{U}}|}$. 
We clarify that if 
\begin{equation}
\label{eq:APP}
\mathbf{A}= \mathbf{P}^{\tilde{\mathcal{B}}}_{\mathcal{U}} \cdot (\mathbf{P}^{\tilde{\mathcal{B}}}_{\mathcal{U}})^\mathrm{T}
\end{equation}
 is achieved, the association result of $\mathbf{A}$ satisfies the transitive consistency. The proof is as following.\\
\textbf{\textit{{Inference.}}} When there is a {{row-concatenation} permutation  matrix} from $\tilde{\mathcal{B}}$ to $\mathcal{U}$, for $\forall \mathcal{B}_i \subseteq \tilde{\mathcal{B}}$ there is also a (injective) mapping to $\mathcal{U}$ {(represented by the permutation matrix $\mathbf{P}^{\mathcal{B}_i}_{\mathcal{U}}$)}, so we can also get an {injective consistency relation} from $\mathcal{U}$ to $\mathcal{B}_i$, i.e. {$\varphi_{\mathcal{U} \mathcal{B}_i}$}. 
%\redcolor{
%	Due to the introduction of $\mathcal{U}$, we can drop the strong assumption in Section~\ref{sec:idea} and prove it for the more general situation. At this time, Eq.~(\ref{eq:phi}) is no longer necessarily holds, and we need to change our thinking.
%	\\
%}
Since $\mathcal{U}$ is the universal set of all appearing subjects {and $\mathcal{B}_i \mapsto \mathcal{U}$ is an injective mapping}, we can get
\begin{equation}
	\label{Eq:mappingtransitive}
	%\varphi_{\mathcal{B}_i \mathcal{B}_j} = \varphi_{\mathcal{U} \mathcal{B}_j} \circ %\varphi_{\mathcal{B}_i \mathcal{U}}  ,           \forall \mathcal{B}_i,\mathcal{B}_j \subseteq %\mathcal{B}.
	\varphi_{\mathcal{B}_i \mathcal{B}_j} = \varphi_{\mathcal{B}_i \mathcal{U}} \triangleleft \varphi_{\mathcal{U} \mathcal{B}_j}  ,           \forall \mathcal{B}_i,\mathcal{B}_j \subseteq \tilde{\mathcal{B}}.
\end{equation}
where the notations are same with those in Section~\ref{sec:idea}.

For any subject ${O}$ in set $\mathcal{B}_1$ and also appearing in other $n-1$ sub-sets in $\tilde{\mathcal{B}}$.
Based on the transitivity property in Eq.~(\ref{eq:Transitivitymap}) in Section~\ref{sec:idea}, we define a transitive operation by circularly passing through all these $n$ sets as
\begin{equation}
	\label{Eq:tc(s1)}
	\Phi_{\mathcal{B}_1} \triangleq 
	%\varphi_{\mathcal{B}_n \mathcal{B}_{1}} \circ \varphi_{\mathcal{B}_{n-1} \mathcal{B}_{n}} %\circ \cdots \circ \varphi_{\mathcal{B}_2 \mathcal{B}_{3}} \circ \varphi_{\mathcal{B}_1 %\mathcal{B}_{2}}.
	\varphi_{\mathcal{B}_1 \mathcal{B}_{2}} \triangleleft \varphi_{\mathcal{B}_2 \mathcal{B}_{3}} \triangleleft \cdots \triangleleft \varphi_{\mathcal{B}_{n-1} \mathcal{B}_{n}} \triangleleft \varphi_{\mathcal{B}_n \mathcal{B}_{1}}.
\end{equation}
Note that, here we do not assume that all the sets share the same persons. Thus, $\Phi_{\mathcal{B}_1}$ is not an identity mapping like Eq.~(\ref{eq:Transitivitymap}).
We take advantage of the universal set $\mathcal{U}$ in Eq.~(\ref{Eq:mappingtransitive}) and get
\begin{equation}
	\Phi_{\mathcal{B}_1} = 
	%\varphi_{\mathcal{U} \mathcal{B}_1} \circ \varphi_{\mathcal{B}_n \mathcal{U}} \circ \cdots %\circ \varphi_{\mathcal{U} \mathcal{B}_2} \circ \varphi_{\mathcal{B}_1 \mathcal{U}}
	%
	\varphi_{\mathcal{B}_1 \mathcal{U}} \triangleleft \varphi_{\mathcal{U} \mathcal{B}_2} \triangleleft \cdots \triangleleft \varphi_{\mathcal{B}_n \mathcal{U}} \triangleleft \varphi_{\mathcal{U} \mathcal{B}_1},
\end{equation}
in which we plug in the variable $O$ and get the result of the transitive association relations as
\begin{equation}
	\Phi_{\mathcal{B}_1}(O)=\varphi_{\mathcal{U} \mathcal{B}_1}(\varphi_{\mathcal{B}_n \mathcal{U}} (\cdots \varphi_{\mathcal{U} \mathcal{B}_2}(\varphi_{\mathcal{B}_1 \mathcal{U}}(O)))).
\end{equation}
On the other hand, since $O$ appears in all $n$ sets, we can easily get
\begin{equation}
	\varphi_{\mathcal{B}_j \mathcal{U}}(\varphi_{\mathcal{U} \mathcal{B}_j}(\varphi_{\mathcal{B}_i \mathcal{U}}(O))) = \varphi_{\mathcal{B}_i \mathcal{U}}(O), \forall \mathcal{B}_i,\mathcal{B}_j \subseteq \tilde{\mathcal{B}}.
\end{equation}
From the above two equations, we can get
\begin{equation}
	\Phi_{\mathcal{B}_1}(O)=\varphi_{\mathcal{U} \mathcal{B}_1}(\varphi_{\mathcal{B}_1 \mathcal{U}}(O))=O.
\end{equation}
So far, the connection of subject $O$ forms a closed loop. $\hfill\blacksquare$

{
Next, we discuss how {to use} a differentiable loss to model~Eq.~\eqref{eq:APP}.
According to the theory of matrix, if $\mathbf{A}$ is a real symmetric matrix (guaranteed by the constraint of symmetric-consistency), Equation~\eqref{eq:APP} is equivalent to $\mathbf{A}$ being $\textbf{positive-semidefinite}$~\cite{huang2013consistent}. 
%Then we can get $\mathrm{rank}(\mathbf{A})=\mathrm{rank}(\mathbf{P}^{\tilde{\mathcal{B}}}_{\mathcal{U}}) \leq \min(|\tilde{\mathcal{B}}|,|\mathcal{U}|)=|\mathcal{U}|$, where $|\mathcal{U}|$ represents the total number of subjects in $\mathcal{U}$. 
%However, we can't know $\mathcal{U}$ in advance. 
%Therefore, similar to previous works~\cite{}, we transform the constraint as $\|\mathbf{A}$ being a $\textbf{low-rank}$ matrix. 
It can be proved that we can use the nuclear norm $\|\mathbf{A} \|_*$ (i.e., the sum of singular values) to approximate the above constraint of $\mathbf{A}$.\\
\textbf{\textit{{Inference.}}} Let $\mathbf{e} =  ({e_1,e_2,...,e_N}) $ denote the eigenvalues of $\mathbf{A}$, from the knowledge of matrix theory we can know that when $\mathbf{A}$ is a real symmetric matrix, its singular value is equal to the absolute value of the eigenvalue, then we can get
\begin{equation}
	\label{eq:nuclear}
	\|\textbf{A}\| _*=\|\mathbf{e}\| _1, %\  \mathrm{rank}(\textbf{Y})=\|\mathbf{e}\| _0,
\end{equation}
which represent the sum of the absolute values of the eigenvalues.% and the number of non-zero eigenvalues, respectively.

	For positive-semidefinite constraint, we want $e_i \geq 0,\forall i \in \{1,2,\cdots ,N\}$, we know $|e_i| \geq e_i$, then we can minimize $|e_i| - e_i,\forall i \in \{1,2,\cdots ,N\}$ to guarantee that, i.e. minimizing $\sum_{i=1}^{N} (|e_i|-e_i)$. Here we have
	\begin{equation}
		\sum_{i=1}^{N} (|e_i|-e_i)=	\sum_{i=1}^{N}|e_i|-\sum_{i=1}^{N}e_i=	\|\textbf{A}\| _*-\mathrm{tr}(\textbf{A}),
	\end{equation}
	with the constraint of $\mathcal{L}_\mathrm{Pse}^\mathrm{M}$, each of the diagonal elements in $\textbf{A}$ is close to 1, i.e. $\mathrm{tr}(\textbf{A})$ is close to $N$. So we can only minimize $\|\mathbf{A} \|_*$ to guarantee positive-semidefinite constraint. $\hfill\blacksquare$
}
%2) For low-rank constraint, we need to minimize the $l_0$ norm, which is an NP-hard nonconvex optimization problem. We usually use its optimal convex approximation $l_1$ norm (i.e., nuclear norm) instead of it. \\ 

This way, we use the nuclear norm to approximately guarantee the transitive-consistency
\begin{equation}
	\label{Eq:core}
	\mathcal{L}_\mathrm{Trs}^\mathrm{M} =  \|\textbf{A} \|_*.
\end{equation}

%Based on the property described in~\cite{C2011Convexity} that the $L_1$ norm of a matrix is the optimal convex approximation of its $L_0$ norm when all the singular values of the matrix are less than $1$,
%the appropriate relaxations of minimizing the nuclear norm can be used to reflect this constraint.
%The singular values and eigenvalues of matrix $\textbf{P}$ are the same because it is symmetric and the eigenvalues of matrix $\textbf{P}$ are less than $1$ according to the constraints in Eq.~(\ref{Eq:0}).

\subsection{The New Association and Tracking Scheme}
\label{sec:endtoend}
%\textbf{Multi-view MHAT Framework.}

%\bluecolor{
%With the above spatial-temporal association network in Section~\ref{sec:network} training on the proposed self-supervision 
%%loss in Section~\ref{sec:loss}} 
{With the above spatial-temporal self-consistency for appearance learning in 
Section~\ref{sec:appfeature} and assignment matrix learning in Section~\ref{sec:assmatrix}}, our method can be trained with 
the videos without tracking and association labels.
The total loss of the whole framework is calculated as 
%\begin{equation}
%	\mathcal{L} = \mathcal{L}_{\mathrm{Sym}}^\mathrm{A} + \mathcal{L}_{\mathrm{Trs}}^\mathrm{A} + %\mathcal{L}_\mathrm{Pse}^\mathrm{M} + \mathcal{L}_\mathrm{Sym}^\mathrm{M} + %\mathcal{L}_\mathrm{Trs}^\mathrm{M} .
%\end{equation}
\begin{equation}
	\label{Eq:totalloss}
	\mathcal{L} =  \mathcal{L}_{\mathrm{Sym}}^\mathrm{A} + \mathcal{L}_{\mathrm{Trs}}^\mathrm{A} + \mathcal{L}_\mathrm{Pse}^\mathrm{M} +  \mathcal{L}_\mathrm{Sym}^\mathrm{M} + \mathcal{L}_\mathrm{Trs}^\mathrm{M}.
\end{equation}
%	where $\alpha_{1},\alpha_{2},\alpha_{3},\alpha_{4},\alpha_{5}$ are preset parameters.

In the inference stage, we propose a new scheme to jointly achieve the 
association and tracking tasks. After getting the output $\mathbf{A}$ of STAN, we partition the matrix according to different time and views, on each of which uses the Hungarian algorithm with matching threshold ${M}$ to get the binary matrix as the final spatial permutation matrices. For temporal association, we use the cascade matching and IOU matching for temporal subject association following the MOT algorithm DeepSort~\cite{bewley2016simple}.
Considering the space occupation and running speed, we use the frames from two consecutive points of time and all views to generate the spatial-temporal affinity matrix as the input of STAN in the training and inference.
The proposed MvMHAT scheme is summarized in Algorithm~\ref{alg:framework}.
Specifically for the human ID assignment strategy, let’s explain it by an example. In view $v_1$, we assume a person $P$ firstly appears at time $t_1$, then disappears at $t_2$, and re-appears at $t_3$. In this case, at $t_1$, we use Algorithm~\ref{alg:framework} to assign a new ID to $P$ and initialize a multi-view tracklet. At $t_2$, we mark the unmatched tracklet to be `sleep' in view $v_1$. Here the tracklet of subject $P$ in $v_1$ interrupts but the multi-view tracklet of $P$ maintains because it still appears in other views. At $t_3$, we use the multi-view subject association results to help match $P$ to the slept tracklet in view $v_1$. For the traditional MOT, it is hard to continuously track $P$ if it disappears for a long time -- $P$ is usually assigned with a new ID when it re-appears. 
%However, as a limitation of our method, incorrect cross-view association results at $t_3$ will cause wrong tracking results.

\begin{algorithm}[h] 
	\small
	\label{alg:framework}
	\caption{MvMHAT framework:}
	\KwIn {$\mathcal{V} = \{\mathcal{V}_i| i = 1, ..., V\}$: a group of synchronized videos captured from different $V$ views.}
	\KwOut{Tracked subject bounding boxes with associated ID.}	
	Split the multi-view videos into $T$ frames, respectively.\\
	%\For{$v=1:S$}{
	%tracker[$v$].tracks=[].
	%}
	\For{$t = 1 : T$ }{
		Detect the $N_t^v$ subjects in frame $t$ for each view $v$: 
		$\mathcal{B}_t^v = \{B_{t_j}^v|j=1, 2, ..., N_t^v\} (v=1, 2, ..., V)$.
		
		%$Emb=[].$
		
		%Generate spatial permutation matrixs $\mathbf{P}^{v, u}_{t}, (v 
		%= 
		%1, 2, ..., S; u = 1, 2, ..., S; v \neq u)$, and 
		%temporal permutation matrixs $\mathbf{P}^v_{t, 
		%	t-1}, (v = 1, 2, ..., S)$.
		%\For{$v=1:V$}{
		%$\mathbf{E}^v_{t-1}$=GetPreFrameMatchTracksFeature(tracker[$v$].tracks).
		%Use the latest features of the previous frame successfully matched trajectories in %tracker[$v$].tracks to generate $\mathbf{E}^v_{t-1}$.
		
		%$\mathbf{E}$.add($\mathbf{E}^v_{t-1}$,$\mathbf{E}^v_t$).
		%$\tilde{\mathbf{E}}$.add($\mathbf{E}^v_{t-1}$).
		%}
		%\For{$v=1:V$}{
		%$\mathbf{E}^v_{t-1}$=GetPreFrameMatchTracksFeature(tracker[$v$].tracks).
		%$\mathbf{E}^v_t = \Phi (\mathcal{B}^v_t)$.
		
		%$\mathbf{E}$.add($\mathbf{E}^v_{t-1}$,$\mathbf{E}^v_t$).
		%$\tilde{\mathbf{E}}$.add($\mathbf{E}^v_{t}$).
		%}
%		\redcolor{According to Eq.~(\ref{Eq:concate}), use the latest features of the previous frame successfully matched trajectories in tracker[$v$].tracks ($\mathbf{E}^v_{t-1}$) and $\Phi (\mathcal{B}^v_t)$ ($\mathbf{E}^v_t$) to generate $\tilde{\mathbf{E}}$.}
		
		%Generate the features $\tilde{\mathbf{E}}$ for all subjects $\mathcal{B}_t^v$ in $\tilde{\mathcal{B}}$.
		
		Generate features $\tilde{\mathbf{E}}$ from the matched trajectories in previous frames and $\mathcal{B}_t^v$ at frame $t$ in each view $v$.
		 
		%$\mathbf{A}=$SplitMatrixAndRowSoftmax($\mathbf{E}  \cdot \mathbf{E}^\text{T}$).
		Generate $\tilde{\mathbf{X}}$ from $\tilde{\mathbf{E}}$ according to Section~\ref{sec:STGA}.
		
		$\mathbf{A}=$STAN($\tilde{\mathbf{X}}$)
		
		$\mathbf{A}=\frac{1}{2}(\mathbf{A}+\mathbf{A}^\text{T})$
		
	Generate spatial permutation matrices $\mathbf{P}^{v, u}_{t,t},\mathbf{P}^{v, u}_{t,t-1}, (v,u = 1, 2, ..., V; v \neq u)$ from $\mathbf{A}$.
		
		\For{$v=1:V$}{
			
		\tcp{temporal association}tracker[$v$].unmatchdets,tracker[$v$].matches=
		DeepSort(tracker[$v$].tracks, $\mathbf{E}^v_t$, $\mathcal{B}_t^v$)
			
			}
		\For{$v=1:V$}{
				%tracker[$v$].unmatchdets,spatialmatches=SpatialMatch(tracker[$1:S$].matches,$\mathbf{P}_{t}$,%tracker[$v$].unmatchdets) \\
				%tracker[$v$].matches=Combine(tracker[$v$].matches,spatialmatches) \\
			\For{$p$ $\in$ $\mathrm{tracker}[$v$]\mathrm{.unmatchdets}$}{
				\tcp{spatial association using the current and previous frames}\If {$(\exists u, q, \mathcal{T}_q, \text{s.t.}
						\mathbf{P}^{v, u}_{t,t}(p, q) = 1 \land (\mathcal{T}_q,q) \in$ $\mathrm{tracker}[$u$]\mathrm{.matches}$$) 
						\vee (\exists u(u \neq v), \mathcal{T}_q, \text{s.t.}\mathbf{P}^{v, u}_{t,t-1}(p,\mathcal{T}_q)=1)$ }{
						tracker[$v$].matches.add(($\mathcal{T}_q$,$p$))
						
						\If{$\mathcal{T}_q \notin$ $\mathrm{tracker}[$v$]\mathrm{.tracks}$}{
							tracker[$v$].tracks.add($\mathcal{T}_q$)
						}
					}
				\Else{
						Initialize new tracklet $\mathcal{T}_p$ with $\mathcal{B}_{t_p}^v$.
						tracker[$v$].tracks.add($\mathcal{T}_p$)
						tracker[$v$].matches.add(($\mathcal{T}_p$ , $p$))
				}
				tracker[$v$].unmatchdets.remove($p$)
			}
			%\For{$p=1:N_t^v$}{
			%	\If{$\exists r, \text{s.t. } \mathbf{P}^{v}_{t, t-1}(p, 
			%		r) = 1 $}
			%	{
			%		Associate tracklet $\mathcal{T}_r$ with 
			%		$\mathcal{B}_{t_p}^v$.
			%	}
			%	\ElseIf {$\exists (u, q), \text{s.t. } u < v \land 
			%		\mathbf{P}^{v, u}_{t}(p, q) = 1$}{
			%		Associate tracklet $\mathcal{T}_q$ with 
			%		$\mathcal{B}_{t_p}^v$.
			%	}
			%	\Else{
			%		Initialize new tracklet $\mathcal{T}_p$ with 
			%		$\mathcal{B}_{t_p}^v$.
			%	}					
			%}	
		}		
	}		
	\Return Bounding boxes with ID numbers in tracks
\end{algorithm}

\subsection{Implementation Details}
%We traverse all the frames from different views along the whole video during the whole training stage. 
%\bluecolor{
%In the inference stage, for temporal association, we use the cascade matching and IOU matching for temporal subject association following the MOT algorithm DeepSort~\cite{bewley2016simple}.
%When constructing the input matrix, in addition to the detection subject features of each view in the current frame, similar to the training stage, the previous frame is also required. Unlike the training stage, here we use the latest features of the successfully matched trajectories from the previous frame in the trajectory library of each view (i.e., the successfully matched detection subject features from the previous frame) to build the input matrix. Thanks to this, in the inference stage, for detection subjects that are not matched by temporal association, we can help match them with spatial association results from the current frame and $\textbf{the previous frame as well}$.
%}

\textbf{Network settings.}
We use annotated detection while training and use results of Detectron~\cite{wu2019detectron2} while inference.
We use ResNet-50~\cite{he2016deep} as the feature extraction backbone network in all experiments, which has outputs of 
1000-d features. 
In Eq.~(\ref{eq:tau}), we set $\epsilon = 0.1$ and $\delta = 0.5$.
{
The parameters ${M}$ in Eq.~(\ref{eq:pseudo1}), $m_{1}$ in Eq.~(\ref{eq:margin}), and $m_{2}$ in Eq.~(\ref{eq:margin2}) are set as $0.5, 0.5$ and $0.4$, respectively. In Eq.~(\ref{Eq:Focalloss}), we set $\gamma=2$ and $\alpha$ is calculated from the ratio of negative samples to total samples. 
%In Eq.~(\ref{Eq:totalloss}), we set $\alpha_{1}=1.0, \alpha_{2}=1.0, \alpha_{3}=5.0, \alpha_{4}=0.002, \alpha_{5}=0.0001$.
}
We use the Pytorch backend for implementing the proposed
network and run it on a computer with RTX 3090 GPU. Our network is 
trained on 8700 groups of frames in MvMHAT training dataset and 8000 groups of frames in MMP-MvMHAT training dataset for 15 epochs with the initial learning rate $10^{-5}$.%\bluecolor{, the inference speed is 30+ FPS}.

{
\textbf{STAN pre-training strategy.}
We also develop a self-supervised pre-training strategy for STAN by automatically generating the training data. i.e., input matrix {$\tilde{\mathbf{X}}$} and ground-truth assignment matrix ${\mathbf{A}^\mathrm{gt}}$. 
Specifically, we simulate the multi-view multi-subject scene and generate the multi-view over-time subject assignment matrix ${\mathbf{A}^\mathrm{gt}}$, on which 
we add the random noise to obtain the input matrix {$\tilde{\mathbf{X}}$}. {We constrain the row sum of each submatrix $\mathbf{X}_{t,s}^{u,v}$ in $\tilde{\mathbf{X}}$ as 1 to simulate the results obtained by the row softmax operation.}
We also define a controllable parameter error rate $e_r$, which considers the probability of occurrence, i.e., the same person in different {frames} show big appearance difference
%\greencolor{(two subjects denoting the same person have low matching scores in $\tilde{\mathbf{X}}$)} 
while the different person show the similar difference.
%\greencolor{(two subjects denoting the different person have high matching scores in $\tilde{\mathbf{X}}$)}.
In the experiments, we perform the end-to-end training using the framework of Fig.~\ref{fig:2} with the pre-train STAN using above strategy and the STAN without pre-training, respectively.
%which controls  of the two cases: 1) when the subjects between two frames have a matching relationship but all values in that row in the input matrix are small; 2) when the subjects between two frames don't have a matching relationship but one value in that row in the input matrix is very large. 
%We generate four datasets based on $e_r$ = 5\%, 10\%, 15\%, and 30\%, each with 10,000 pairs of matrices for training and 2,000 pairs of matrices for testing. We use Eq.~(\ref{Eq:SymC}),~(\ref{Eq:core}) and use $\hat{\mathbf{Y}}$ instead of ${\mathbf{A}_\mathrm{sudo}}$ in ~(\ref{Eq:Focalloss}) to pretrain STAN. 
%%The matrix is consistent with the setup of the training and inference stage, and consists of two consecutive time and multiple views subject. 
%%first define that the total number of people in the scene is 9$\sim$15 and the number of view is 3$\sim$6, 60\%$\sim$100\% of them randomly appear in each view at the first time point. At the second time point, 0$\sim$2 person walk out randomly from each view, and 0$\sim$2 people randomly walk in from the scene. We generate a list of all subject ground truth ids corresponding to the order of Eq.~(\ref{Eq:concate}), which can be extended to $\hat{\mathbf{Y}}$ easily. 
%For , we . In order to simulate the real input situation, we need to ensure that the row sum of each sub-matrix is 1 after the noise matrix is divided by time and view. 
}

\section{MvMHAT Benchmark}
\label{sec:benchmark}
%\textbf{A new synthetic dataset MvMhT-syn}
\subsection{Datasets}

\subsubsection{A new MvMHAT dataset}
\textbf{Dataset Collection. }To the best of our knowledge, most previous MOT datasets with multiple views covering an overlapped region are 
relatively small and only used for algorithm testing. 
For both training and testing the proposed framework, we build an exclusive large-scale video dataset -- MvMHAT benchmark, for multi-view 
multi-human association and tracking task. For reducing the cost of data collection and annotation while 
maintaining the usefulness and credibility of the proposed dataset, some videos and corresponding annotations in MvMHAT are from two available datasets, 
i.e., Campus~\cite{xu2016multi-view}, EPFL~\cite{2008Multicamera}.
Besides, we have also collected 12 video groups containing 46 sequences, with each group has three to four views. 
To enrich the way of data collection, different from previous videos captured by fixed cameras, these videos are collected with four wearable 
cameras, i.e., GoPro, which cover an overlapped area present with multiple people and are from significantly different directions, e.g., 
near 90-degree view-angle difference. We then manually synchronize them and annotate the bounding box and the ID for each subject on all $30,900$ frames.
%This way, we collect four synchronous videos, from which we extract the 
%same $2,860$ frames from the four videos, with a maximum number of $10$ 
%persons in each frame.

\begin{table}[htbp] 
%	\vspace{-8pt}
	\small
	\caption{{Data source and statistics of MvMHAT {dataset}.}} \label{tab:dataset}
	\vspace{-10pt}
	% \footnotesize
	\begin{center}
		\begin{tabular}{l c c c c c}
			\Xhline{1pt}
			Source   &Campus  &EPFL &Self Collected &Total\\
			\hline
			\# Group   &6  &8 &12 &26  \\ 
			\# Sequence &22 &30 &46 &98\\
			\# View & 3-4  &3-4 & 3-4 &3-4\\			
			Avg. Length &1500 &900 &672  &928\\
			Avg. Subject &14 &8 &10 &10\\
			Sum. Length &33,000 &27,000 &30,900 &90,900\\
			\Xhline{1pt}  
		\end{tabular}
	\end{center}
%	\vspace{-10pt}
\end{table}

%CVMHT~\cite{Han2019Complementary}, view : 3 video : 2  +  view : 2 
%video 
%: 3

\textbf{Dataset statistics and splitting.} 
As shown in Table~\ref{tab:dataset}, in total, the dataset in MvMHAT contains 26 multi-view video groups with 98 (single-view) sequences.
Each video group contains several temporal-synchronized videos with multiple views, e.g., 3 -- 4 views.
The average length of each sequence is 928 frames, and average ten subjects appear in each video.
The dataset, in total, contains over 90 thousand frames. We further split the dataset into training and testing datasets, each of 
which contains 13 video groups, and the ratio of containing frames in the training and the testing datasets is about 2:1.
To guarantee the diversity of the testing data, the testing videos contain the videos from Campus, EPFL, and Self collected. 
%	We will release this dataset to the public.

\subsubsection{MMP-MvMHAT dataset}
{
\textbf{Dataset description.}
A recent challenge has collected a new dataset namely Multi-camera Multiple People Tracking (MMPTRACK)~\cite{han2023mmptrack}, which aims to tackle multiple people tracking using multiple RGB cameras. 
It provides long continuous videos that can be used to track people in relatively closed, tight, and complex scenes for long periods of time. 
All videos are recorded by multiple cameras at different locations in an indoor scene, e.g., retail, lobby, industry, cafe and office (each scene has 4-6 cameras), where each camera has the overlapping field of view (FOV) with at least one of the other cameras.
%There are five simulated indoor scenes in the dataset, 
Each scene is divided into four half-day sessions for recording (two sessions for training, one session for validation, and one session for testing).
% with seven completely different people participating in each session and the same group of people in different scenes in the same session performing unscripted spontaneous performance based on the environment setting. 
In total, this dataset has 28 people with different genders, ages and guarantees the training, {validation} and testing sets don't have the same people. 
Due to environments are crowded and cluttered, the occlusion of people occurs frequently, and in some scenes, such as industry, people wear the same work clothes and have similar appearance difference, which makes this dataset very challenging.

\textbf{Dataset reconstitution. }
MMPTRACK dataset provides two sub-tracking challenges: 1) top-down view tracking; 2) single-view tracking separately, both of which are different from our simultaneous association and tracking task. 
This way, based on the original MMPTRACK dataset, we reconstruct a new MMP-MvMHAT dataset and include it for evaluating our task. 
In order to ensure that the MMP-MvMHAT and MvMHAT datasets have approximate data scale, we choose the first 2,000 frames of the four scenes of the lobby, industry, cafe and office videos from all two half-day sessions in the original training set to build ours (i.e. total 8,000 groups of frames for training). Since the origin testing set does not provide ground truth annotation and we need to use our own metrics for evaluation, we use the original validation set to build our test set. 
We use the first 1,000 frames of the four scenes mentioned above from one half-day session in the original validation set as our testing set (i.e. total 4,000 frames for testing). 
In MMP-MvMHAT dataset we use the annotated subject detection boxes for training and testing in the experiments. Note that, we do not use the calibration information of the cameras to fit the more general multi-view camera settings.
}

\subsection{Evaluation metrics} 
%The metrics contains two aspects:
%We evaluate the performance of our method from two aspects, including 
%the over-time multiple human tracking (MHT) and cross-view multi-human 
%association (MHA). 
%	\\
\textbf{MHT metrics.}
We first apply the commonly used MOT metric, i.e., multiple object tracking accuracy (MOTA) 
proposed in~\cite{Bernardin2008Evaluating} for the single-view tracking performance evaluating as in MOT Challenge~\cite{Lealtaix2015MOTChallenge}.
A key task of the MvMHAT task is to associate and track the same subject along the time. We are more concerned about the ID related metrics~\cite{2016Performance} - ID precision~(IDP), ID recall~(IDR), and ID F$_1$ measure~(IDF$_1$) in evaluation. {We also include a new but effective metric -- High Order Tracking Accuracy (HOTA)~\cite{luiten2020hota}, which explicitly balances the effects of performing accurate detection, association and localization into a single unified metric to evaluate the MOT performance.}
\\
\textbf{{MvMHA} metrics.} We further apply the metrics for cross-view association evaluation, i.e., AIDP, 
AIDR, and AIDF$_1$, by expanding the cross-view association metrics in~\cite{Han2019Complementary,han2021multiple}. Specifically, AIDP and AIDR denote the multi-view subject association precision and recall, respectively. Given the subject IDs in all views, we take two views each time and 
compute the pairwise subject matching performance as in~\cite{Han2019Complementary,han2021multiple}, whose average on all 
views is used as a multi-view association metric. Based on AIDP and AIDR, the association F$_1$ score is computed as $\mathrm{AIDF_1}=\frac{2 \times \mathrm{AIDP \times AIDR}}{\mathrm{AIDP}+\mathrm{AIDR}}$. Following~\cite{Bernardin2008Evaluating,han2021multiple}, we also apply 
multi-view multi-human association accuracy %\begin{equation}
$
\mathrm{MHAA}=1-\frac{\sum_{t}({\mathrm{MS}_t}+{\mathrm{FP}_t}+2{\mathrm{MM}_t})}{\sum_{t}{{N}_t}},$
%\end{equation}
where $\mathrm{MS}_t$, $\mathrm{FP}_t$, $\mathrm{MM}_t$ are the false 
negative, false positive, and mismatched pairs at frame $t$. ${N}_t$ is 
the total number of subjects within all views at time $t$. 
The MHAA follows the definition and calculation pattern of MOTA.
\\
\textbf{{MvMHAT} metrics.}
For the overall performance evaluation of MvMHAT problem, we first take a 
simple average and get the MvMHAT \textit{F$_1$ score}, i.e., $\mathcal{F}$
= mean(IDF$_1$, AIDF$_1$) and MvMHAT \textit{accuracy score}, i.e., 
$\mathcal{A}$ = mean(MOTA, MHAA). 
{We also define a new metric STMA (Spatial-Temporal Matching Accuracy) for MvMHAT evaluation. 
	Given a time window with the length $t$,  we first generate the ground-truth matching and the corresponding predicted spatial-temporal subject matching matrix defined in Eq.~(\ref{Eq:combineX}), which include both the cross-view and over-time subject association result across all views for consecutive $t$ frames. We use the human bounding box overlap to establish the subject correspondence between the two matrices, we also use zero padding on the FP (FN) subject matching results in ground truth (predicted) matching matrix to make them the same size.	
	Then we calculate the F$_1$ score between them. 
	By sliding the time window with the step of $\frac{t}{2}$ through the whole video, F1 is calculated for each time window and averaged to obtain the final result as STMA score.  We set $t=5, 10, 30$ and get $\mathcal{S}$@5, $\mathcal{S}$@10 and $\mathcal{S}$@30 to evaluate the MvMHAT performance of the tracker for short, medium and medium-long time periods.}
%	for the test video starting from the first frame, using the subjects for $t$ consecutive time points and , respectively
  \begin{table*}[htbp] 
  	\vspace{-0.cm}
  	\small
  	\caption{Comparative results of different methods on the proposed 
  		MvMHAT benchmark. 
  		%			IDP$\uparrow$, IDR$\uparrow$ , 
  		%			IDF$_1$$\uparrow$, 
  		%			MOTP$\uparrow$ , MOTA$\uparrow$  are standard MOT 
  		%			metrics 
  		%			for over time tracking. AIDP$\uparrow$ , AIDR$\uparrow$ , 
  		%			AIDF$_1$$\uparrow$ , MHAA$\uparrow$ are the metrics 
  		%			for evaluating the cross-view association. $\mathcal{A}$$\uparrow$, 
  		%			$\mathcal{F}$$\uparrow$  are the overall metrics for evaluating 
  		%			the MVMHAT task.(\%)
  	}%\vspace{-0.3cm}
  	\label{tab:res_all}
  	\vspace{-10pt}
  	\centering
  	\begin{center}
  		\begin{tabular}[c]{lccccc|cccc|ccccc}
  			\Xhline{1pt}   
  			\multirow{2}{20pt}{Method} &\multicolumn{5}{c}{Over-Time 
  				Tracking} &\multicolumn{4}{c}{Cross-View Association} 
  			&\multicolumn{5}{c}{Overall} 
  			\\\cline{2-6} \cline{7-10} 	\cline{11-15}
  			&IDP   &IDR  &IDF$_1$    &MOTA &HOTA &AP  &AR  &AF$_1$ 
  			&MHAA&$\mathcal{A}$&$\mathcal{F}$&$\mathcal{S}$@5&$\mathcal{S}$@10&$\mathcal{S}$@30\\ 
  			\hline%\multirow{7}{5pt}{\rotatebox{90}{Baselines}} &FairMOT 
  			%\\ 
  			{Tracktor++~\cite{2019Tracking}} &54.2&40.1&46.1&66.5&42.8&34.3&14.6&	20.5&37.1&51.8&33.3 &44.9 &	44.2 &	42.5 
  			\\
  			\rowcolor{mygray}
  			{CenterTrack~\cite{zhou2020tracking}} &44.3 &33.5 &	38.1  &	63.5 &	37.8 	&29.7 &	9.1 &	13.9& 	34.1 &	48.8 &	26.0 &42.3 &	41.4 &	38.9 
  			\\
  			%DMAN &63.2 &60.0 &61.6 &838 &2025 &79.3 &67.1 &52.4 &26.4 
  			%&35.1 &226588 &41.4 \\
  			{TraDeS~\cite{wu2021track}} &46.7 &	43.2& 	44.9 &	69.5 &	42.9 &	32.4 	&14.0 &	19.6 &	36.0 &	52.8 &	32.2 &48.5 	&47.8 &	45.8 
  			\\
  			\rowcolor{mygray}
  			{TrackFormer~\cite{meinhardt2022trackformer}} &52.3 	&47.2 	&49.6  	&70.4 &	47.3 &47.8 &	23.2 &	31.3 &	40.2 &	55.3 &	40.4 &	54.2 	&53.6 &	51.9 \\

  			{CenterTrack~} (P) &43.8 &	33.7 &	38.1 	&63.1 &	37.7 &	31.9 &	9.3 &	14.4 &	34.3 &	48.7 &	26.3& 42.8 &	41.9 	&39.4 
  			\\
  			%\rowcolor{mygray} 
  			\rowcolor{mygray}
  			{TraDeS~ (P)} &53.9 &	50.5 &	52.1 &	\textbf{70.8} &	46.8 	&38.7 &	19.7 &	26.1 &	38.5 &	54.7 &	39.1 &51.5 &	50.9 &	49.2 \\
		
  			 \hline
  			%\rowcolor{mygray}
  			%DeepCC~\cite{Ristani2018Features} &40.8 &24.6 &30.7 &82.2 
  			%&47.5 &11.2 &2.9 &4.6 &30.4&39.0&17.6\\
  			{DeepCC~\cite{Ristani2018Features}} &44.7 &	44.2 &	44.4&	63.9 &	41.1 &	57.9 &	34.8 	&43.4 &	43.8 &	53.9 &	43.9 &56.6 &	55.6 &	52.8 
  			\\
  			%CVMHT~\cite{Han2019Complementary} &51.1 &36.2 &42.4 &82.3 
  			%&54.1 &32.8 &26.4 &29.2 &41.7&47.9&35.8\\ 
  			\rowcolor{mygray}
  			{SVT~\cite{3Dposetracking}} &47.9 &	47.2 &	47.6  &	65.4 &	43.1 &	61.7 &	45.7 &	52.5 	&50.4 &	57.9 &	50.0 &61.4 	&60.5 &	58.4 
  			\\
  			{Prior~\cite{gan2021self}} &53.1 	&52.0 &52.5 &64.7 &47.9 	&53.0 	&46.4 &49.5 &51.7 &58.2 &51.0 &60.4 &	59.5 &	58.1 
  			\\
  			\hline \hline
  			Ours  &\textbf{58.5} &	\textbf{57.4} &	\textbf{57.9} &	66.3 &	\textbf{51.8} &	\textbf{63.8} &	\textbf{56.0} &	\textbf{59.6} 	&\textbf{59.7} &	\textbf{63.0} &	\textbf{58.8} &\textbf{67.1} &	\textbf{66.3} &	\textbf{65.0} 
  			\\
  			\Xhline{1pt}  
  		\end{tabular}
  	\end{center} 
  	%	\vspace{-10pt}
  \end{table*}

%and multi-object matching accuracy (mvMOMA).
%
%to comprehensively evaluate the MHAT performance, following the 
%traditional MOT metrics 
%in~\cite{Bernardin2008Evaluating,2016Performance},
%
%
%%\end{equation}
%We then define three metrics, i.e., {mvMS}, {mvFP}, {mvMM} to calculate 
%the numbers of missed matches, false positives, and mismatch pairs for 
%multi-view subject association. 
%Specifically, for $n$ views with in total $N$ subjects, we first get 
%ground-truth and predicted matching matrix with the dimension of $N 
%\times N$ for all views.
%An example is shown in Fig.~\ref{fig:metric}. The metric mvMS counts 
%the 
%number of missed ground-truth matching pairs, 
%the mvFP counts the number of falsely detected matching pairs, and the 
%mvMM metric counts the number of mismatches. 
\section{Experimental Results}
\label{sec:experment}
\subsection{Comparison Results}

\label{sec:results}
%We evaluate the proposed method on the MvMHAT benchmark to demonstrate 
%the rationality of the problem, and compare it with other approaches to 
%verify the effectiveness of our method. 

\subsubsection{Baseline methods} As discussed above, we actually did not find 
existing methods that can directly handle our MvMHAT problem for 
comparison. Therefore, we try to include sufficient related approaches 
with some modifications for comparison.
\\
$ \bullet $ We first select {four} state-of-the-art MOT methods for 
single-view video, including CenterTrack~\cite{zhou2020tracking}, 
Tracktor++~\cite{2019Tracking}, TraDeS~\cite{wu2021track} and {TrackFormer~\cite{meinhardt2022trackformer}} for 
comparison. {We know that the single-view MOT methods only handle the tracking in 
	each 
	video but not including the cross-view association.
	For comparison, we additionally help them by providing the ground-truth 
	unified IDs for the subjects among different views when they appear in 
	each video for the first time.
	The over-time tracking on each video can propagate the IDs to subsequent 
	frames, which we can use to associate the subjects across views and over 
	time.  }\\
%\bluecolor{
%	$ \bullet $
%	We also include a multi-target multi-camera tracking (MTMCT) method 
%	DeepCC~\cite{Ristani2018Features} for comparison.
%	Note, DeepCC is used to handle the human tracking and re-identification 
%	(re-id) using multiple cameras covering different areas.
%	We modify it to handle the proposed MvMHAT following the deep re-id 
%	features 
%	and BIP based correlation clustering method used in DeepCC.   \\}
{
	$ \bullet $
	We also include two methods on multi-camera multi-human tracking for comparison, i.e., DeepCC~\cite{Ristani2018Features} and SVT~\cite{3Dposetracking}.
%Specifically for DeepCC, we use an off-the-shelf person re-identification method~\cite{zhong2018camera} trained on the Market-1501 dataset~\cite{zheng2015scalable} as the feature extraction network to extract appearance features of each human bounding box. 
%Then as originally we construct $\tilde{\mathbf{X}}$ using two consecutive points of time and all views subjects. As in ~\cite{Ristani2018Features}, we use the spectral clustering to generate spatial-temporal permutation matrices $\mathbf{P}$ as the MvMHAT result, where we additionally provide the ground truth number of persons contained in each $\tilde{\mathbf{X}}$ as clusters number. 
Note that, DeepCC is used to handle the human tracking and re-identification (re-id) using multiple cameras covering different areas. We modify it to handle the proposed MvMHAT following its basic idea of using deep re-id features and clustering algorithm. Specifically, we use an off-the-shelf person re-identification method~\cite{zhong2018camera} trained on the Market-1501 dataset~\cite{zheng2015scalable} as the feature extraction network to extract the appearance feature of each human bounding box. 
Then as originally we construct $\tilde{\mathbf{X}}$ using two consecutive points of time and all views subjects. We use the spectral clustering to generate spatial-temporal permutation matrices $\mathbf{P}$ as the MvMHAT result, where we additionally provide the ground truth number of persons contained in each $\tilde{\mathbf{X}}$ as clusters number. 
For SVT, we solve $\mathbf{P}$ for the symmetric and cycle consistency constraints using the Augmented Lagrangian Method (ALM) algorithm as in~\cite{3Dposetracking}. We also modified these two methods based on the inference framework in Algorithm~\ref{alg:framework} as our method for fair comparison. 
\\
}
%\bluecolor{
%	$ \bullet $ Besides, we also take the approach CVMHT 
%	in~\cite{Han2019Complementary} as a comparison method, which takes 
%	{pairwise-view} videos as input and cannot directly handle our problem 
%	with multiple ($\geq$ 2) views.
%	We divide the video groups in our dataset into pairwise video pairs and 
%	evaluate CVMHT on each pair, respectively.\\}
{
	$ \bullet $ Besides, we also compared with the prior version of this work in~\cite{gan2021self}, i.e. a self-supervised method that does not consider the dummy nodes and the spatial-temporal global information in the assignment matrix generation.
}

For a fair comparison, we use the same human detection provided by the 
commonly used detector~\cite{wu2019detectron2} {in MvMHAT dataset and use annotation detections in MMP-MvMHAT dataset}, for all the comparison 
methods and the proposed method. 
We also reserve the results provided by {CenterTrack} and TraDeS using the 
private detector {denoted as (P)}, as shown in Tables~\ref{tab:res_all} {and~\ref{tab:mmp}}.
We clarify that all the networks in the comparison methods we use are 
the public version trained on the original training dataset.
{For relatively fair, we use the trained models of the comparison methods and do not re-train them on our datasets, since all these methods need supervision with labeled data, which is not used in our	self-supervised method.}

\begin{table*}[htbp] 
	\vspace{-0.cm}
	\small
	\caption{Comparative results of different methods on MMP-MvMHAT benchmark. 
		%			IDP$\uparrow$, IDR$\uparrow$ , 
		%			IDF$_1$$\uparrow$, 
		%			MOTP$\uparrow$ , MOTA$\uparrow$  are standard MOT 
		%			metrics 
		%			for over time tracking. AIDP$\uparrow$ , AIDR$\uparrow$ , 
		%			AIDF$_1$$\uparrow$ , MHAA$\uparrow$ are the metrics 
		%			for evaluating the cross-view association. $\mathcal{A}$$\uparrow$, 
		%			$\mathcal{F}$$\uparrow$  are the overall metrics for evaluating 
		%			the MVMHAT task.(\%)
	}%\vspace{-0.3cm}
	\label{tab:mmp}
	\vspace{-10pt}
	\centering
	\begin{center}
		\begin{tabular}[c]{lccccc|cccc|ccccc}
			\Xhline{1pt}   
			\multirow{2}{20pt}{Method} &\multicolumn{5}{c}{Over-Time 
				Tracking} &\multicolumn{4}{c}{Cross-View Association} 
			&\multicolumn{5}{c}{Overall} \\\cline{2-6} \cline{7-10} 
			\cline{11-15}
			&IDP   &IDR  &IDF$_1$   &MOTA &HOTA &AP  &AR  &AF$_1$ 
			&MHAA &$\mathcal{A}$ &$\mathcal{F}$ &$\mathcal{S}$@5 &$\mathcal{S}$@10 &$\mathcal{S}$@30\\ 
			\hline%\multirow{7}{5pt}{\rotatebox{90}{Baselines}} &FairMOT 
			%\\ 
			{Tracktor++~\cite{2019Tracking}} &67.0 &	56.0 &	61.0 	&	67.2 &	46.2 	&62.0 &	23.2 &	33.8 &	19.1 &	43.1 	&47.4 &45.2 &	44.7 &	43.8 
			\\
			\rowcolor{mygray}
			{CenterTrack~\cite{zhou2020tracking}} &35.9 	&24.1 &	28.8  &	48.2 &	27.1 &	29.3 &	3.9 &	6.8 	&3.1 &	25.7 &	17.8& 26.0 &	25.0 &	23.0 
			\\
			%DMAN &63.2 &60.0 &61.6 &838 &2025 &79.3 &67.1 &52.4 &26.4 
			%&35.1 &226588 &41.4 \\
			{TraDeS~\cite{wu2021track} }&59.7 &	50.1 	&54.5  &	66.1 &	42.7 &	54.5 &	17.3 &	26.2 &	13.3 	&39.7 &	40.4 &41.5 &	41.0 &	40.2 
			\\
			\rowcolor{mygray}
			{TrackFormer~\cite{meinhardt2022trackformer}} &41.8 &	28.6 	&34.0 	&	46.7 &	30.2 	&39.9 &	5.3 &	9.4 &	3.2 &	25.0 &	21.7 	&27.2 &	26.6 &	25.1 \\
			
			{CenterTrack} (P)&31.9 &	24.1 	&27.4&	39.6 &	26.1 &	32.6 &	3.7 &	6.7 &	3.1 	&21.3 &	17.1 &25.7 	&24.8 &	22.8 
			\\
			\rowcolor{mygray}
			%\rowcolor{mygray} 
			{TraDeS (P)} &57.3 &	49.8 &	53.3 	 &	63.0 &	42.0 &	53.1 &	17.1 &	25.9 &	13.2 &	38.1 	&39.6 &41.0 &	40.5 &	39.7 
			\\

			\hline
			%{w/o $\mathcal{L}_\mathrm{Trs}^\mathrm{M}$} &65.4 &	65.9 	%&65.7 &	96.4 &	94.9 &	68.6 &	54.8 &	37.1 &	44.3 &	%33.6 	&64.3 	&55.0 \\
			%\hline
			%\rowcolor{mygray}
			%DeepCC~\cite{Ristani2018Features} &40.8 &24.6 &30.7 &82.2 
			%&47.5 &11.2 &2.9 &4.6 &30.4&39.0&17.6\\
			{DeepCC~\cite{Ristani2018Features}} &51.6 &	52.5 &	52.1 &	92.5 	&59.3 &	42.7 &	23.4 &	30.2 &	19.7 &	56.1 &	41.2 &47.3 &	46.4 &	43.1 
			\\
			%CVMHT~\cite{Han2019Complementary} &51.1 &36.2 &42.4 &82.3 
			%&54.1 &32.8 &26.4 &29.2 &41.7&47.9&35.8\\ 
			\rowcolor{mygray}
			{SVT~\cite{3Dposetracking}}&63.1 &	63.4 &	63.3 &	\textbf{96.7}  &	68.8 &	53.8 	&33.4 &	41.2 &	29.8 	&63.1 &	52.3 &55.6 &	55.0 &	53.3 
			\\
			{Prior~\cite{gan2021self}} &58.6 &	58.8 &	58.7  &	93.7 	&65.0 &	35.4 &	21.2 &	26.5 &	20.3 &	57.0 	&42.6 &44.7 &	44.0 &	41.8 
			\\
			\hline \hline
			Ours&\textbf{67.1} 	&\textbf{67.6} 	&\textbf{67.3}  &	95.0 &	\textbf{70.2} &	\textbf{62.1} 	&\textbf{42.5} &	\textbf{50.4} &	\textbf{40.6} &	\textbf{67.8} &	\textbf{58.9} &\textbf{61.8} &	\textbf{61.2} &	\textbf{59.3} 
			\\
			\Xhline{1pt}  
		\end{tabular}
	\end{center} 
	%	\vspace{-10pt}
\end{table*} 

\subsubsection{Results on MvMHAT dataset}
Table~\ref{tab:res_all} shows the comparative results of our methods 
with 
the baseline methods.
For the single-view MOT methods, i.e., Tracktor++, CenterTrack, TraDes {and TrackFormer}, we first evaluate the over-time tracking performance using the 
standard MOT metrics.
We can see that the state-of-the-art approach TraDes with the private 
detector provide the best MOTA score among all competitors.
Note that, MOTA mainly focuses on the object detection accuracy 
and precision during tracking~\cite{luiten2020hota}. On the contrary, 
the 
ID-based metrics, i.e., IDP, IDR and IDF$_1$ evaluate the ID association 
and consistency over time. This paper is more concerned about the 
latter. 
We can see that the proposed method outperforms all the above methods in 
IDF$_1$ score. {The composite metric HOTA also demonstrates the effectiveness of our method for over-time tracking.}
We then show the cross-view association results in the middle of 
Table~\ref{tab:res_all}.
%{Note that, for the single-view MOT methods, we additionally help them by providing the ground-truth 
%	unified IDs for the subjects among different views when they appear in 
%	each video for the first time. }
From the first {six} rows, we can see that the cross-view 
association performances provided by the single-view MOT methods are 
poor {even with the help of providing the ground-truth unified IDs for the subjects among different views for the first time.}
This is because, without the cross-view re-associating mechanism during 
tracking, the association will fail once occurring the subject ID switch.

%\bluecolor{
%	For the multi-view tracking approaches, i.e., DeepCC and CVMHT, which 
%	also perform not well enough particularly for the cross-view association.
%	This is because the DeepCC is mainly used for long-term trajectory 
%	re-identification but not good at synchronously associating the subjects 
%	across views simultaneously.
%	For CVMHT~\cite{Han2019Complementary}, it emphasizes the spatial 
%	distribution feature but simplifies the appearance for subject matching, 
%	which is not very suitable in our setting.}

{
	For the first two multi-view tracking approaches, i.e., DeepCC and SVT, which have significantly better ability to associate across view, but are still worse than our method. This is because our method is an end-to-end training framework that not only generates assignment results using spatial-temporal global consistency information, but also use it to train the feature extraction network simultaneously, while DeepCC and SVT are two-stage methods and consider spatial-temporal association only in the assignment phase. 
	Compared to the prior version of our method, the results are further improved. This is because we consider the more general and frequent dummy nodes cases caused by the occlusion or out of view, and we also take advantage of the spatial-temporal global consistency information in the assignment phase.
}
For the overall performance metrics, i.e., $\mathcal{A}$, $\mathcal{F}$ and $\mathcal{S}$, on all of 
which our method get the best performance. This significantly verify the effectiveness of the proposed method for MvMHAT problem.

\subsubsection{Results on MMP-MvMHAT dataset}
{
	Table~\ref{tab:mmp} shows the comparative results of our methods 
	with the baseline methods. Since MMP-MvMHAT dataset is a simulated indoor scene dataset, the subjects in it have simpler motion patterns. In some scenes, such as the office scene, many subjects even do not walk around, but just work in their seats. So compared to Table~\ref{tab:res_all}, we can see that most of the methods have higher IDF$_1$ score in over-time tracking except CenterTrack {and TrackFormer}. For the single-view MOT methods Tracktor++ and TraDes, with the over-time tracking becomes more accurate, the corresponding cross-view association AIDF$_1$ scores are also improved. However, for CenterTrack, since the method only uses the subjects of two adjacent frames for association and does not use the appearance features of the subjects, in crowded and complex scenes, i.e., when the subjects are close with frequent mutual occlusions, Centertrack  often generates ID switch in the over-time tracking, resulting in poor MOT results. We also note that TrackFormer performs not very well on MMP-MvMHAT, especially for MOTA. 
	TrackFormer is a algorithm that implements both human detection and association. This way, even if we provide the unified human detection results, it only filters its own detection results based on them. Thus the inaccurate detection result of TrackFormer on MMP-MvMHAT dataset makes the tracking results unsatisfactory. Their poor MOT results also lead to poor cross-view association results accordingly.}
	
	For the multi-view tracking approaches, we are able to find that the performance of cross-view association of these methods decreases compared to Table~\ref{tab:res_all}, because the humans in this dataset have similar appearance, and the human bounding boxes always contain cluttered backgrounds or other objects. This will interfere with the effectiveness of extracted subject features, making the {approach} using appearance similarity for association is incompetent on this dataset. We can also find that our `prior' method does not work well on this dataset, 
	%because it ignores the problem of null relation due to occlusion or out of view, which often happens in dense and crowded scenes. This also illustrates the effectiveness of our improvements made in this work, even in complex and crowded scenes our method still works well.
	because it does not use the spatial-temporal global consistency information when cross-view association, but only uses the similarity of subject appearance between pairwise views to obtain association results, while the appearance feature is no longer reliable for this dataset as discussed above. This also illustrates the effectiveness of our improvements made in this work, even in complex and crowded scenes our method still works well.
	For the overall performance metrics, i.e., $\mathcal{A}$, $\mathcal{F}$ and $\mathcal{S}$, on all of 	which our method performs best among all methods.
}

\begin{table*}[htbp] 
	\vspace{-0.cm}
	\small
	\caption{Ablation study of different variations of our method on MvMHAT benchmark.} 
	\vspace{-10pt}
	\label{tab:ablation}
	\centering
	\begin{center}
		\begin{tabular}[c]{lccccc|cccc|ccccc}
			\Xhline{1pt} 
			\multirow{2}{20pt}{Method} &\multicolumn{5}{c}{Over-Time 
				Tracking} &\multicolumn{4}{c}{Cross-View Association} 
			&\multicolumn{5}{c}{Overall} \\
			\cline{2-6} \cline{7-10} \cline{11-15}
			&IDP   &IDR  &IDF$_1$    &MOTA &HOTA &AP  &AR  &AF$_1$ 
			&MHAA&$\mathcal{A}$&$\mathcal{F}$&$\mathcal{S}$@5&$\mathcal{S}$@10&$\mathcal{S}$@30\\ \hline
			%{w/o Train} &58.8 &57.3 &58.0 &858 &2175 &79.3 &67.8 &6.7 
			%&2.0 &3.1 &321310 &29.6\\
			%{w/o Training} &34.2 &34.6 &34.4 &78.3 &57.2 &23.3 &15.1 
			%&18.3 &27.7&42.5&26.4\\
			{w/o Training} &32.9 &	33.4 &	33.2  &	56.7 &	33.2 &	21.6 &	14.6 	&17.4 &	27.6 &	42.1 &	25.3 &	37.3 &	35.7 &	30.8 
			\\
			\rowcolor{mygray}
			{w/o $\mathcal{L}_{\mathrm{Sym}}^\mathrm{A}$} &56.4& 	55.5 	&55.9 		&65.3 &	50.5 &	62.2 &	55.4 &	58.6 &	58.1 	&61.7 &	57.3 &66.1 &	65.2 	&63.7 
			\\
			
			{w/o $\mathcal{L}_{\mathrm{Trs}}^\mathrm{A}$}& 54.8 &	53.8 	&54.3  &	65.2 &	50.0 &	61.6 &	55.7 &	58.5 	&58.4 	&61.8 &	56.4 &66.2 &	65.3 	&63.8 
			\\
			\rowcolor{mygray}
			{w/o dummy nodes} &56.2 &	55.2 &	55.7  &	66.4 	&49.3 &	57.8 &	50.8 &	54.1 &	56.2 &	61.3 &	54.9 &64.1 	&63.3 &	61.9 
			\\
			
			%{w random pse label}& 55.1 	&54.4 &	54.8 	&64.6 	&50.2 &	58.1 &	%52.8 &	55.4 &	55.0 &	59.8 &	55.1 &63.1 &	62.2 &	60.4 
			%\\
			{w/o $\mathcal{L}_\mathrm{Sym}^\mathrm{M}$} &57.5 	&56.4 	&57.0 	&66.4 	&51.5 &	60.5 &	52.0 &	55.9 &	57.2 	&61.8 &	56.5& 65.0 &	64.2 	&63.0 
			\\
			\rowcolor{mygray}
			{w/o $\mathcal{L}_\mathrm{Trs}^\mathrm{M}$} &58.1 &	57.0 	&57.5 	&66.4 	&52.4 	&61.1 &	54.3 &	57.5 &	57.9 	&62.1 &	57.5& 66.0 &	65.2 	&63.9 
			\\
			\hline
			
			{w/o Association} &64.9 &	\textbf{63.2} &	\textbf{64.0}  &	\textbf{67.9} &	\textbf{55.3} &	42.9 &	3.5 	&6.5 &	30.3 &	49.1 &	35.2 &44.7 	&43.9 &	42.7 
			\\		
			\rowcolor{mygray}
			{w/o Tracking} &\textbf{65.8} &	45.8 &	54.0  &	47.8 &	46.1 &	\textbf{68.2} &	52.0 	&59.0 &	56.4 &	52.1 &	56.5 &59.2 	&58.2 &	56.7 
			\\		
			%{w/o $\mathcal{L}_{\mathrm{SSIM}}$}  &52.7 &51.7 &52.2 &79.2 
			%&64.5 &56.4 &41.7 &48.0 &49.2&56.9&50.1\\
			%{w/o $\mathcal{L}_{\mathrm{TSIM}}$} &49.4 &48.6 &49.0 &79.2 
			%&65.3 &54.7 &36.1 &43.5 &46.2&55.8&46.2\\\hline
			%\rowcolor{mygray}
			%{w/o Association} &63.2 &61.3 &62.2 &79.4 &67.7 &22.3 &5.1 
			%&8.3 &30.1&48.9&35.2\\
			%{w/o Track} &51.1 &50.2 &50.7 &6183 &6210 &79.0 &55.3 &53.7 
			%&48.5 &51.0 &83364 &47.9\\
			%{w/o Tracking} &59.1 &42.0 &49.1 &79.8 &42.5 &55.8 &44.4 
			%&49.4 &47.6&45.0&49.2\\ \hline
			%			w random sample &52.2 &51.3 &51.8 &79.2 &64.2 
			%&50.6 &32.1 &39.3 &42.9&53.5&45.5\\
			%\rowcolor{mygray}
			%{w/o Relax} &48.1 &47.4 &47.8 &79.1 &64.1 &43.9 &24.9 &31.8 
			%&39.2&51.6&39.8\\
			%{w/o Margin} &31.2 &29.6 &30.4 &79.0 &61.0 &24.2 &11.0 &15.2 
			%&29.0&45.0&22.8\\			
			%{w/o temporature} &48.8 &47.9 &48.3 &2609 &2649 &79.1 &65.4 
			%&16.2 &5.3 &8.0 &255628 &28.9\\
			%\rowcolor{mygray}
			%{w/o Temperature} &38.0 &38.1 &38.1 &78.6 &59.7 &16.6 &9.6 
			%&12.2 &26.6&43.2&25.1\\
			\hline \hline
			Ours &58.5 &	57.4 &	57.9  &	66.3 &	51.8 &	63.8 &	\textbf{56.0} &	\textbf{59.6} 	&\textbf{59.7} &	\textbf{63.0} &	\textbf{58.8} &\textbf{67.1} &	\textbf{66.3} &	\textbf{65.0} 
			\\		
			\Xhline{1pt}  
		\end{tabular}
	\end{center} %\vspace{-0.3cm}
\end{table*}

\begin{table*}[htbp] 
	\vspace{-0.cm}
	\small
	%\caption{Ablation study of different values of error rate $e_r$ on the %proposed MvMHAT benchmark.} 
	\caption{Ablation study of STAN {with different settings} on MvMHAT benchmark.}
	\vspace{-10pt}
	\label{tab:ablationinite}
	\centering
	\begin{center}
		\begin{tabular}[c]{lccccc|cccc|ccccc}
			\Xhline{1pt} 
			\multirow{2}{20pt}{Method} &\multicolumn{5}{c}{Over-Time 
				Tracking} &\multicolumn{4}{c}{Cross-View Association} 
			&\multicolumn{5}{c}{Overall} \\\cline{2-6} \cline{7-10} 
			\cline{11-15}
			&IDP   &IDR  &IDF$_1$   &MOTA &HOTA &AP  &AR  &AF$_1$ 
			&MHAA&$\mathcal{A}$&$\mathcal{F}$&$\mathcal{S}$@5&$\mathcal{S}$@10&$\mathcal{S}$@30\\ \hline
			%{w/o Train} &58.8 &57.3 &58.0 &858 &2175 &79.3 &67.8 &6.7 
			%&2.0 &3.1 &321310 &29.6\\
			%{w/o Training} &34.2 &34.6 &34.4 &78.3 &57.2 &23.3 &15.1 
			%&18.3 &27.7&42.5&26.4\\
			{w/o STAN} &57.5 &	56.4 &	56.9  &	66.0 &	50.9 &	\textbf{64.0} 	&53.3 &	58.2 &	57.3 &	61.6 &	57.5 &66.0 &	65.2 &	63.9 
			\\
			\rowcolor{mygray}
			{w one time point} &58.3 	&57.1 &	57.7  &	66.3 	&51.3 &	63.9 &	52.2 &	57.5 &	57.0 &	61.6 &	57.6 &65.8 &	65.0 &	63.9 
			\\ \hline
			{unpretrain STAN} &57.3 &	56.2 &	56.8 &	66.1 	&51.5 &	59.3 &	53.2 &	56.1 &	57.3 &	61.7 &	56.4 &64.7 	&63.9 	&62.5 
			\\
			\rowcolor{mygray}
			{$e_r=0.05$} &56.4 	&55.3 	&55.8  &	66.3 &	51.0 	&61.5 &	55.8 &	58.5 &	58.6 &	62.5 &	57.2 &66.2 	&65.4 	&64.0 
			\\
			
			{$e_r=0.15$} &58.1 	&57.0 	&57.5&	65.9 &	\textbf{52.3} 	&61.4 &	55.9 &	58.5 &	58.7 &	62.3 &	58.0 &66.3 &	65.5 &	64.2 
			\\
			\rowcolor{mygray}
			{$e_r=0.30$} &56.9 &	55.8 &	56.4 &	66.2 &	51.1 	&60.5 &	54.8 &	57.5 &	58.1 &	62.1 &	56.9 &66.1 	&65.3 	&63.9 
			\\
			
			%{w/o $\mathcal{L}_{\mathrm{SSIM}}$}  &52.7 &51.7 &52.2 &79.2 
			%&64.5 &56.4 &41.7 &48.0 &49.2&56.9&50.1\\
			%{w/o $\mathcal{L}_{\mathrm{TSIM}}$} &49.4 &48.6 &49.0 &79.2 
			%&65.3 &54.7 &36.1 &43.5 &46.2&55.8&46.2\\\hline
			%\rowcolor{mygray}
			%{w/o Association} &63.2 &61.3 &62.2 &79.4 &67.7 &22.3 &5.1 
			%&8.3 &30.1&48.9&35.2\\
			%{w/o Track} &51.1 &50.2 &50.7 &6183 &6210 &79.0 &55.3 &53.7 
			%&48.5 &51.0 &83364 &47.9\\
			%{w/o Tracking} &59.1 &42.0 &49.1 &79.8 &42.5 &55.8 &44.4 
			%&49.4 &47.6&45.0&49.2\\ \hline
			%			w random sample &52.2 &51.3 &51.8 &79.2 &64.2 
			%&50.6 &32.1 &39.3 &42.9&53.5&45.5\\
			%\rowcolor{mygray}
			%{w/o Relax} &48.1 &47.4 &47.8 &79.1 &64.1 &43.9 &24.9 &31.8 
			%&39.2&51.6&39.8\\
			%{w/o Margin} &31.2 &29.6 &30.4 &79.0 &61.0 &24.2 &11.0 &15.2 
			%&29.0&45.0&22.8\\			
			%{w/o temporature} &48.8 &47.9 &48.3 &2609 &2649 &79.1 &65.4 
			%&16.2 &5.3 &8.0 &255628 &28.9\\
			%\rowcolor{mygray}
			%{w/o Temperature} &38.0 &38.1 &38.1 &78.6 &59.7 &16.6 &9.6 
			%&12.2 &26.6&43.2&25.1\\
			\hline \hline
			{Ours ($e_r=0.10$)} &\textbf{58.5} &	\textbf{57.4} &	\textbf{57.9}  &	\textbf{66.3} &	51.8 &	63.8 &	\textbf{56.0} &	\textbf{59.6} 	&\textbf{59.7} &	\textbf{63.0} &	\textbf{58.8} &\textbf{67.1} &	\textbf{66.3} &	\textbf{65.0} 
			\\		
			\Xhline{1pt}  
		\end{tabular}
	\end{center} %\vspace{-0.3cm}
\end{table*}

\subsection{Ablation Study}
%	\textbf{Ablative methods.} To explore the influence of different modules 
%	in our method, we conduct a series of ablative variations.\\
%\subsubsection{Setup}

%\bluecolor{
%	$ \bullet $ w/o Training:
%	We directly use the Resnet-50~\cite{he2016deep} network pre-trained on 
%	ImageNet~\cite{deng2009imagenet} to extract features instead of the one 
%	using the self-supervised training in our method.\\
%	$ \bullet $ w/o $\mathcal{L}_{\mathrm{SSIM}}$:
%	Remove the SSIM loss $\mathcal{L}_{\mathrm{SSIM}}$ in 
%	Eq.~(\ref{eq:loss-s}).\\
%	$ \bullet $ w/o $\mathcal{L}_{\mathrm{TSIM}}$:
%	Remove the TSIM loss $\mathcal{L}_{\mathrm{SSIM}}$ in 
%	Eq.~(\ref{eq:loss-t}).	\\ 
%	$ \bullet $ w/o Association:
%	Remove the association module during tracking, and generate the 
%	association results when the object first appears.\\
%	$ \bullet $ w/o Track:
%	Remove the collaboratively tracking module, and implement the tracking 
%	on 
%	one view to obtain the subject ID in other views by cross-view 
%	association.\\
%	$ \bullet $ w/o. Relax:
%	Replace the relaxed margin with a strict one namely set $m = 1$ in 
%	Eq.~(\ref{eq:margin}).\\
%	$ \bullet $ w/o Margin:
%	Remove the margin namely set $m = 0$ in Eq.~(\ref{eq:margin}).\\
%	$ \bullet $ w/o Temperature:
%	Remove the temperature mechanism namely we set $\tau = 1$ in 
%	Eq.~(\ref{eq:softmax}).\\
%}

\subsubsection{Effectiveness of self-supervised loss} 
We first investigate the effectiveness of self-supervised loss by considering the following ablation studies. \\
{
$ \bullet $ w/o Training:
 We directly use the Resnet-50 pre-trained on
 ImageNet~\cite{deng2009imagenet} and STAN pre-trained on synthetic dataset (with $e_r=10\%$) without using the self-supervised training proposed in our method.\\}
$ \bullet $ w/o $\mathcal{L}_{\mathrm{Sym}}^\mathrm{A}$:
Remove the symmetric-consistency loss for appearance feature learning $\mathcal{L}_{\mathrm{Sym}}^\mathrm{A}$ in Eq.~(\ref{eq:loss-s}).\\
$ \bullet $ w/o $\mathcal{L}_{\mathrm{Trs}}^\mathrm{A}$:
Remove the transitive-consistency loss for appearance feature learning $\mathcal{L}_{\mathrm{Trs}}^\mathrm{A}$ in Eq.~(\ref{eq:loss-t}).\\
$ \bullet $ w/o dummy nodes:
Use a diagonal identity matrix to supervise {$\mathbf{I}$} with a relaxation~\cite{gan2021self} instead of Eq.~(\ref{eq:l1l2}).\\
%$ \bullet $ w random pseudo label:
%When generating binary pseudo label using $\tilde{\mathbf{X}}$ in  Eq.~(\ref{Eq:combineX}), values less than or equal to 0.3 are set to 0, values greater than or equal to 0.7 are set to 1, and the rest are randomly taken as 0 or 1.\\
$ \bullet $ w/o $\mathcal{L}_{\mathrm{Sym}}^\mathrm{M}$:
Remove the symmetric-consistency loss for assignment matrix learning $\mathcal{L}_{\mathrm{Sym}}^\mathrm{M}$ in Eq.~(\ref{Eq:SymC}).\\
$ \bullet $ w/o $\mathcal{L}_{\mathrm{Trs}}^\mathrm{M}$:
Remove the transitive-consistency loss for assignment matrix learning $\mathcal{L}_{\mathrm{Trs}}^\mathrm{M}$ in Eq.~(\ref{Eq:core}).
%\bluecolor{
%	As shown in Table~\ref{tab:ablation}, we can see that '\textit{w/o 
%		Training}' has a poor performance, which shows the challenge of the 
%	MvMHAT task and the importance of proposed self-supervised training. Our 
%	loss functions make the network have ability to unsupervisedly find the 
%	potential characteristics of data. '\textit{w/o 
%		$\mathcal{L}_{\mathrm{SSIM}}$}' and '\textit{w/o 
%		$\mathcal{L}_{\mathrm{TSIM}}$}' shows the effectiveness of 
%	$\mathcal{L}_{\mathrm{TSIM}}$ loss function and 
%	$\mathcal{L}_{\mathrm{SSIM}}$ function. $\mathcal{L}_{\mathrm{SSIM}}$ 
%	only considers the relationship between pairwise frames, which leads to 
%	the drop of performance. Similarly, $\mathcal{L}_{\mathrm{TSIM}}$ 
%	focus on considering global information also does not perform as well as 
%	that combining both of 
%	them.\\}

{
	As shown in Table~\ref{tab:ablation}, {we can see that '\textit{w/o Training}' has a poor performance, which shows
		the challenge of the MvMHAT task and the importance of proposed
		self-supervised training. Our loss functions enable the network to
		unsupervised find the potential characteristics of data.} '\textit{w/o $\mathcal{L}_{\mathrm{Sym}}^\mathrm{A}$}' and '\textit{w/o $\mathcal{L}_{\mathrm{Trs}}^\mathrm{A}$}' make over-time tracking results drop significantly, this means we can't have one without the other, in which $\mathcal{L}_{\mathrm{Sym}}^\mathrm{A}$ only considers the relationship between pairwise frames, while $\mathcal{L}_{\mathrm{Trs}}^\mathrm{A}$ focusing on considering transitive-consistency information. For cross-view association, we still have $\mathcal{L}_{\mathrm{Sym}}^\mathrm{M}$ and $\mathcal{L}_{\mathrm{Trs}}^\mathrm{M}$ losses that also guarantee symmetric-consistency and transitive-consistency, so its results drop little.
	'\textit{w/o dummy nodes}' ignores the frequently occurring occlusion or out of view cases, thus leading to the deviation in appearance feature learning with the constrained losses, and the inaccurate feature representation also leads to the degradation of assignment results since matrix $\mathbf{X}$ also provides the pseudo label for training STAN.
%	'\textit{w random pse label}' simply turns $\tilde{\mathbf{X}}$ into a binary matrix without considering the one-to-one matching constraint, it does not give basic assignment results for assignment matrix learning, and also affects appearance feature learning since the framework is end-to-end.\\
	'\textit{w/o $\mathcal{L}_{\mathrm{Sym}}^\mathrm{M}$}' and '\textit{w/o $\mathcal{L}_{\mathrm{Trs}}^\mathrm{M}$}' lead to a significant decrease in cross-view association results, which means that only combining both of them can make better performance. While for over-time tracking, i.e., relying mainly on subject appearance features, they still have $\mathcal{L}_{\mathrm{Sym}}^\mathrm{A}$ and $\mathcal{L}_{\mathrm{Trs}}^\mathrm{A}$ to guarantee symmetric-consistency and transitive-consistency, so the result drop is not very significant.
}

\subsubsection{Effectiveness of association and tracking schema} 
We then investigate the effectiveness of some components in the inference stage by considering the following setting. \\
$ \bullet $ w/o Association:
Remove association module during tracking, and generate the 
association results when the {subject} first appears.\\
$ \bullet $ w/o Tracking:
Remove the collaboratively tracking module, and implement the tracking 
on one view to obtain the subject ID, in other views by cross-view association.

We can see from Table~\ref{tab:ablation} that `\textit{w/o Association}' provides very promising tracking results. It can be explained that, without considering the cross-view association, the method under `\textit{w/o Association}' can avoid more ID switches during tracking. This can be regarded as an upper bound of our method only for tracking, which, however, naturally generates a poor association performance. From comparing the results of `\textit{w/o Tracking}' and `\textit{Ours}', we are surprised to find that with the aggregation of tracking, our method not only improves the performance of over-time tracking but also the cross-view association. This inspires us the temporal tracking can be used as a favor for association. Overall, the integration of collaborative tracking \& association mechanism generates the best results.

\begin{figure*}[h]
	%		\vspace*{-0.1cm}
	\centering	
	\subfigure{
		\begin{minipage}[b]{0.775\textwidth}
			\includegraphics[width=\textwidth]{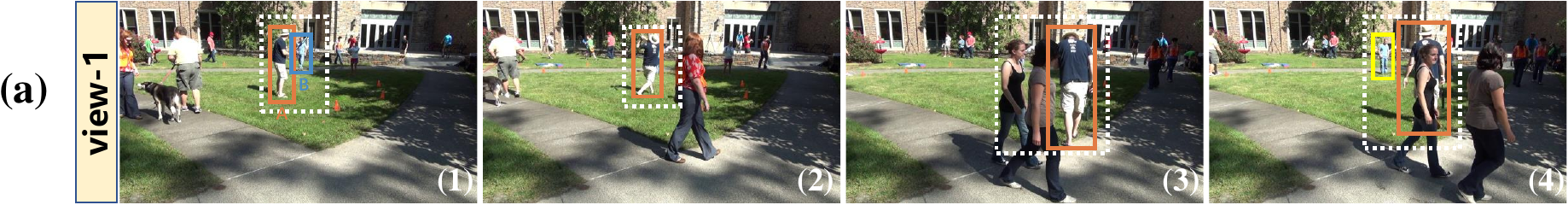} \\
						\vspace*{-0.1cm}\\
			\includegraphics[width=\textwidth]{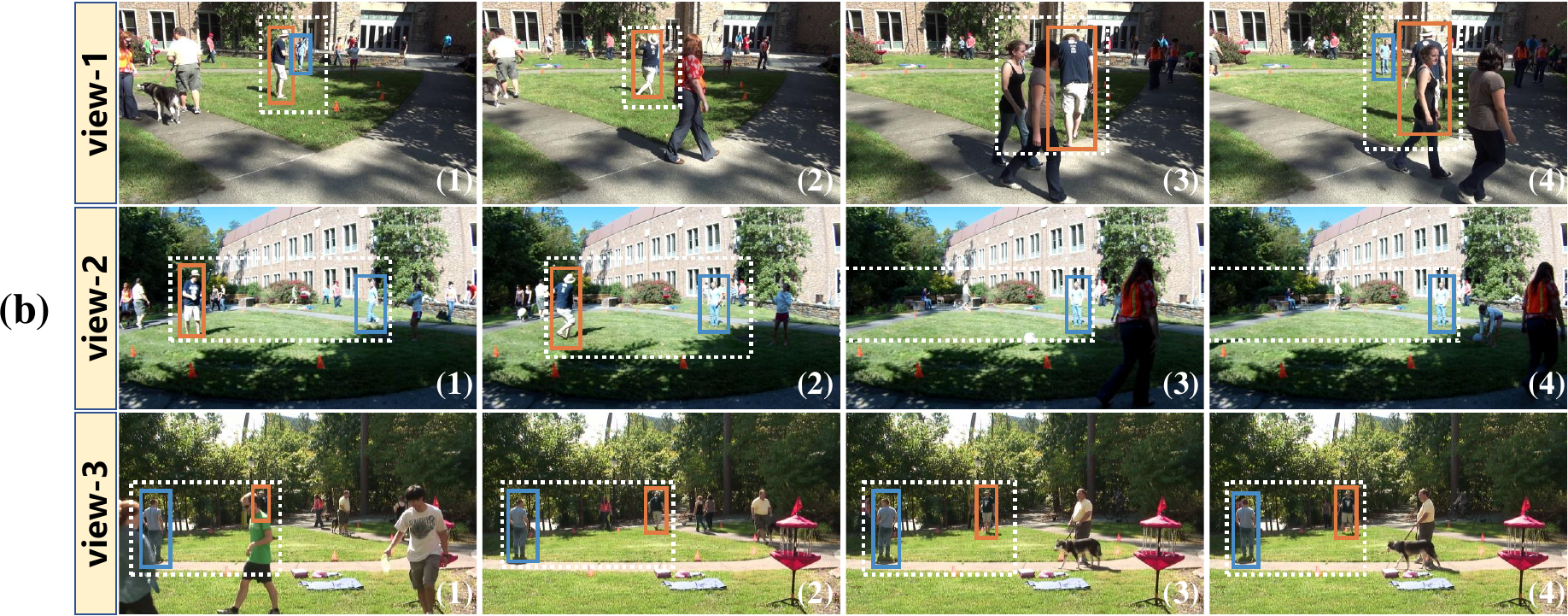} \\
						\vspace*{-0.1cm}\\
			\includegraphics[width=\textwidth]{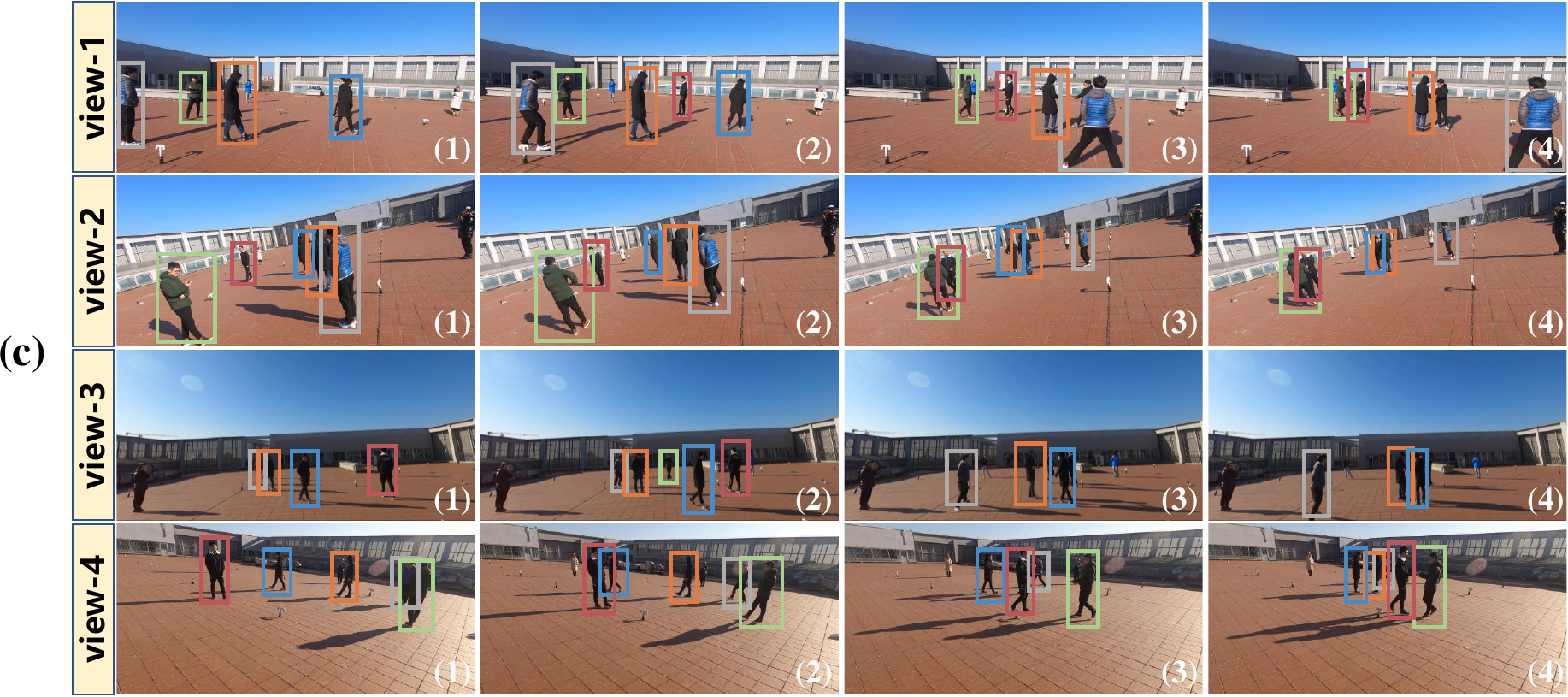}
		\end{minipage}
	}
	
	\caption{An illustration of the qualitative results. {Figure (a) shows the tracking results of the MOT method Tracktor++ for one view, while figure (b) shows the association and tracking results of our method for multiple views. Figure (c) shows the results of our method on a group of multi-view videos in MvMHAT dataset. All the results we show are four sampled frames, which at the same column in each sub-figure are from the same time.}}
	\label{fig:4}
	
\end{figure*}

\subsubsection{Ablation studies on STAN} 
We also in detail investigate the influence of the proposed STAN and its different variations by considering the following settings.\\
$ \bullet $ w/o STAN:
{In the inference stage, }instead of using the output of STAN to generate spatial permutation matrices $\mathbf{P}$, $\mathbf{X}$ is used directly to generate $\mathbf{P}$.\\
$ \bullet $ w one time point:
{In the inference stage, } we use the subject bounding boxes in all views at only one point of time (ignoring the over-time information) to generate spatial permutation matrices $\mathbf{P}$.\\
$ \bullet $ unpretrain STAN:
Do not pre-train STAN, initialize parameters with orthogonal matrix.\\
$ \bullet $ $e_r=0.05,0.15,0.30,0.10$:
Generate pre-training datasets using different error rates $e_r=5\%, 15\%, 30\%, 10\%$ and STAN is pre-trained using these datasets respectively.
%\bluecolor{
%	The results generated by '\textit{w/o Relax}' and '\textit{w/o Margin}' verify the explanation in Section~\ref{sec:loss}. In '\textit{w/o Relax}', we assume $\mathbf{I}$ in Eq.~(\ref{eq:margin}) is identity matrix. However, since frequent occlusion and out-of-view of people, frames from different views or different points of time hardly share the exact same people. In this case, an over-constrained $\mathbf{I}$ tends to give the unmatched people incorrect matchings. On the contrary, '\textit{w/o Margin}' means the strongest relaxation of $\mathbf{I}$, but seems to give a too weak penalty for incorrect matchings.Finally, in a real-world scenario, the spatial-temporal distribution of people is always ruleless, which leads to the different number of people in different views at different time. However, softmax has different soften abilities when the input size changes, whcih makes these output values are affected by the number of people. This way, the \textit{temperature} in {softmax} is beneficial.\\}

{
	First, '\textit{w/o STAN}' instead of using STAN network, it only uses appearance feature learning module to extract the subject features and calculate $\mathbf{X}$ between two frames, then generates the assignment matrix without considering the spatial-temporal global information. We can first see from Table~\ref{tab:ablationinite} that the drop of the result verifies the effectiveness of the assignment matrix learning module. Compared with prior version of our method in Table~\ref{tab:res_all}, we can also find that as an end-to-end framework $\mathcal{L}_{\mathrm{Sym}}^\mathrm{M}$ and $\mathcal{L}_{\mathrm{Trs}}^\mathrm{M}$ losses can also improve the performance of the appearance feature learning module.
	We can see that '\textit{w one time point}'  provides a decreased performance. This is because it considers only spatial consistency without considering temporal consistency when generating the assignment matrix $\mathbf{P}$. Also, it does not use the help of the previous frame when spatial association causing a degradation in the results of cross-view association.
}

{
	As shown in Table~\ref{tab:ablationinite}, we can see from '\textit{unpretrain STAN}' that even if STAN is trained from scratch, it can still achieve satisfactory results. This is because $\mathcal{L}_{\mathrm{Sym}}^\mathrm{A}$ and $\mathcal{L}_{\mathrm{Trs}}^\mathrm{A}$  losses can ensure the correctness of the appearance feature learning module, and therefore guaranteeing the correctness of the pseudo label generated by $\tilde{\mathbf{X}}$, providing STAN network with the correct optimizing direction. 
	As an end-to-end network, {assignment matrix learning module} can also improve the performance of the appearance feature learning module, forming a virtuous circle. Therefore, the self-supervision method proposed in this work has strong practical application, that is, STAN is not depended on the pre-training.
	For the pre-trained STAN, we can find that most of the results are better than the unpre-trained STAN, and the results have the best over-time tracking and cross-view association ability when the error rate of the generated dataset $e_r=10\%$. 
	We therefore use this pre-trained STAN model as the initial state of our method on the MvMHAT dataset and MMP-MvMHAT dataset.
	Note that, as discussed above, the data for the pre-training of STAN are self-generated without any annotation cost, which does not break the self-supervised manner of this work.
}

\subsection{Qualitative Evaluation}
Figure~\ref{fig:4}(a) shows a long-term tracking result of Tracktor++. 
We can see that two players are practicing baseball in the white dotted 
area in (1). However, because of the frequent moving of the player A in 
(2), and comings and goings of pedestrians in (3), the player B is 
occluded frequently. 
Thus, the single view MOT methods have to always try to match B with all 
the tracklets, which leads to lots of incorrect ID switches, e.g., in 
(4).
As shown in Fig.~\ref{fig:4}(b), our method leverages the complementary 
characteristic of multiple views,  and we ensure that any people can be 
observed in at least one view at all times.
The integration of multi-view tracklets makes good use of both temporal 
and spatial information. 
The proposed MvMHAT scheme, to some extent, can help track the 
re-appearing detections 
that cannot be matched by the tracking module, while the continual 
tracking can help 
associate the subjects that cannot be matched by the association module.
This way, a spatial-temporal connection is established for all observed 
people, which \textit{has the potential to better handle the 
	long-term 
	tracking}. 
In Fig.~\ref{fig:4}(c), people are walking and interacting with each 
other with various activities. Based on the results by MvMHAT, we can 
capture 
the details of every people from all-around perspectives.
This demonstrates the \textit{potential of the proposed MvMHAT for board 
	applications}, e.g., sports games, {indoor/}outdoor surveillance, {outdoor law enforcement}, which aim to 
capture both global and local details of involved people.

{
	\section{Discussion}
	\label{sec:discussion}
	\subsection{Analysis of Self-supervised Framework}
%	As shown in Fig.~\ref{fig:2}, our method uses an end-to-end self-supervised training framework to train two networks simultaneously: feature extraction network and spatial-temporal assignment network. 
	In the training phase, we use $n$ frames from different time and views as an input batch. 
	For the feature extraction network, as discussed in Section~\ref{sec:loss}, $\textbf{I}_{\mathrm{S}}$ and $\textbf{I}_{\mathrm{T}}$ matrices are calculated for any pairwise and triplewise frames among $n$ frames according to Eqs.~(\ref{eq:Is}) and~(\ref{eq:It}). That is, for a batch of n frames, we generate {${\mathrm{P}(n,2)+\mathrm{P}(n,3)}$} matrices for self-supervised learning{, where $\mathrm{P}(n,m)$ represents the number of permutations in which $m$ elements are taken from $n$ different elements.}
%	For training the STAN, we use all detection subjects in $n$ frames to construct $\tilde{\mathbf{X}}$ matrix to be constraint by the self-supervised losses. 
%	Through the above analysis, it is obvious that the larger $n$ is, the more constraints there are for the network, and the smaller solution space of feature extraction network and spatial-temporal assignment network parameters learning will be, that is, the subject features extracted and the assignment matrix generated by the network have to maintain symmetric-consistency and transitive-consistency over more frames.
	Considering the space occupation, we use the frames from two consecutive points of time and all views as an batch in our method. 
	Even so, for four views, about 400 $\mathbf{I}$ matrices are generated in each batch, which are  constrained by the proposed self-supervised losses $\mathcal{L}_{\mathrm{Sym}}^\mathrm{A}$ and $\mathcal{L}_{\mathrm{Trs}}^\mathrm{A}$  in Eqs.~(\ref{eq:loss-s}) and~(\ref{eq:loss-t}).
	{The feature extraction network is also constrained by $\mathcal{L}_\mathrm{Pse}^\mathrm{M}$, $\mathcal{L}_\mathrm{Sym}^\mathrm{M}$ and $\mathcal{L}_\mathrm{Trs}^\mathrm{M}$ of STAN in Eqs.~(\ref{Eq:Focalloss}),~(\ref{Eq:SymC}),~(\ref{Eq:core}) as an end-to-end framework.}
%	the losses of  do not only constrain the learning of spatial-temporal assignment network, but also further constrain the learning of feature extraction network. 
	Therefore, the proposed fully self-supervised learning framework can generate enough self-constraints effectively and be able to produce promising results.
	\subsection{Limitation Analysis}
	%\textbf{Field-of-view (FOV) overlap of the multi-view. }
	In this paper, we consider the case of inconsistent involved subjects in different views/time. However, when the subjects overlap in different views is too small, i.e., only a very small fraction of the subjects are the same people in each view, the proposed method does not work well. 
	We can see this through the $\mathcal{L}_\mathrm{Trs}^\mathrm{M}$ loss in Eqs.~(\ref{Eq:core}), i.e., the nuclear norm loss. 
	The nuclear norm is usually used as a convex approximation of the matrix rank, so we also require the output $\mathbf{A}$ of STAN to be a low-rank matrix by minimizing the nuclear norm. 
	However, when the overlap across view is very small, {i.e., $|{\mathcal{U}}|$ is very close to $|\tilde{\mathcal{B}}|$,} the ground truth of the spatial-temporal assignment matrix $\mathbf{A}$ is no longer low-rank. {Meanwhile, when the FOV overlap of cameras is very small, the same subject may not appear in multiple cameras at the same time, and cross-view association becomes less important accordingly. 
	This problem is more like the existing MTMCT (multi-target multi-camera tracking) task where the cameras are without FOV overlap. 
	Therefore, we clarify that the proposed MvMHAT problem and the existing MTMCT task \textit{are complementary to each other}, the study on both of which can promote the development of the multi-camera MOT task.
%	As the difference between MvMHAT and  is described in the related work, this problem becomes a cross-view human Re-ID problem instead of a multi-graph matching problem.
}
	
	%\textbf{Number of subjects. }
	The performance of our method is also related to the number of subjects in the scene and gets worse when there are too many or too few people in each view video. When there are too many subjects, dense crowds can lead to inaccurate subject detection and appearance feature representation results. When there are too few subjects, for the extreme case, when there is only one subject in frame $j$, the $\mathbf{X}_{ij}$ matrix obtained from the row softmax has only one column and each matching score in it is 1, which cannot correctly reflect the matching relation of the subjects between frame $i$ and frame $j$.
}

\section{{Conclusion And Future Work}}
\label{sec:conclusion}
%In this paper, we have studied a relatively new problem -- MvMHAT, which 
%is different from the existing MOT problem and has promising development 
%potential.
%For fully excavating the peculiarity MvMHAT, we model this problem as a 
%self-supervised learning task and propose an end-to-end framework to 
%handle it.
%To promote the study on this new topic, we have also built a new MvMHAT 
%benchmark for performance evaluation. Experimental results verify the 
%rationality of our problem formulation, the usefulness of the proposed 
%benchmark and the effectiveness of our method.
%In the future, we hope to develop more research and further attract more 
%interests in the community on this topic.
In this paper, we have studied a relatively new problem -- MvMHAT, which 
is different from the existing MOT problem and has promising development 
potential.
For fully excavating the peculiarity of MvMHAT, based on the rationale of spatial-temporal self-consistency, we model this problem as a self-supervised learning task and propose an end-to-end framework to handle it. 
To promote the study on this new topic, we have also built a couple of MvMHAT 
benchmarks for network training and performance evaluation. 
Experimental results verify the rationality of our problem formulation, the usefulness of the proposed 
benchmark, and the effectiveness of our method.
%In the future, we hope to develop more research and further attract more 
%interests in the community on this topic.

{For future work, we hope to develop more human features to better handle this problem. We mainly explore the appearance consistency of the subjects over time and across views in this work. The input matrix of STAN, i.e., the affinity matrix $\tilde{\mathbf{X}}$, is also constructed based on appearance similarity. However, through the comparison between Tables~\ref{tab:res_all} and \ref{tab:mmp}, we can see that the cross-view association result of our method on MMP-MvMHAT dataset is lower than that of MvMHAT dataset. 
In other words, when the subjects' appearances are similar and the backgrounds are cluttered, our method can not play a good role when only using subject appearance. 
In fact, the self-consistency constraint proposed in this paper is the consistency of matching relation, not just the consistent appearance similarity. 
Therefore, in the future, more human features can be explored, such as the consistency of subject pose across views~\cite{Ass4Tracking} and the consistency of subject motion pattern over time~\cite{wang2021track}. We can use the proposed self-supervised loss for their consistency learning, so as to better conduct the multi-view multi-human association and tracking. 
This paper takes the first step to achieve the MvMHAT task using a fully self-supervised method to obtain a promising result. In the future, we hope to develop more research and further attract more interests in the community on this topic.}

\bibliographystyle{IEEEtran}
\bibliography{IEEEabrv,MHA,CVMOT,MvMHAT}

\end{document}